%% file: main.tex
\pdfoutput=1
\documentclass[11pt]{article}
\PassOptionsToPackage{hyperfootnotes=false}{hyperref}
\usepackage[preprint]{acl}
\usepackage{times}
\usepackage{latexsym}
\usepackage[T1]{fontenc}
\usepackage[utf8]{inputenc}
\usepackage{microtype}
\usepackage{courier}
\usepackage{graphicx}
\usepackage{amsmath,amsfonts,amssymb}
\usepackage{array,multirow,makecell,tabularx,booktabs}
\usepackage{xurl}
\usepackage{ragged2e}
\usepackage{enumitem}
\usepackage{placeins}
\usepackage{needspace}
\usepackage[most]{tcolorbox}
\usepackage{listings}
\renewcommand{\cite}{\unskip\nobreakspace\citep}

\providecommand{\Description}[2][]{}
\newcommand{\PaperFigure}[3]{%
  \includegraphics[width=#1,keepaspectratio]{#2}%
}
\renewcommand{\theHfigure}{main.\arabic{figure}}
\renewcommand{\theHtable}{main.\arabic{table}}
\input{supplementary_style}

\hypersetup{
  pdftitle={LongSearchEval: Evaluating Long-Horizon Web Search Capabilities of Large Language Models via a Trajectory-Aware Agent Framework with Failure-Node Localization},
  pdfauthor={Jialiang Liu; Shusheng Li; Wenjun Tan}
}
\title{LongSearchEval: Evaluating Long-Horizon Web Search Capabilities of Large Language Models via a Trajectory-Aware Agent Framework with Failure-Node Localization}
\author{
\normalsize\bfseries
Jialiang Liu\textsuperscript{1}\footnotemark[2]\qquad
Shusheng Li\textsuperscript{2,3,4}%
\thanks{Corresponding authors.}%
\thanks{These authors contributed equally to this work.}\qquad
Wenjun Tan\textsuperscript{2,3,4}\footnotemark[1]
\\[0.2em]
\small\normalfont
\textsuperscript{1}College of Information Science and Engineering,
Northeastern University, China\\[-2.4pt]
\small\normalfont
\textsuperscript{2}Key Laboratory of Intelligent Computing of Medical Images,
Ministry of Education,\\[-2.4pt]
\small\normalfont
Northeastern University, China.\\[-2.4pt]
\small\normalfont
\textsuperscript{3}School of Computer Science and Engineering,
Northeastern University, Shenyang 110819, China\\[-2.4pt]
\small\normalfont
\textsuperscript{4}National Frontiers Science Center for Industrial Intelligence
and Systems Optimization,\\[-2.4pt]
\small\normalfont
Northeastern University, China\\[-1pt]
\small\normalfont
\texttt{tanwenjun@cse.neu.edu.cn}
}

\begin{document}
\maketitle
\begingroup
\urlstyle{rm}
\Urlmuskip=0mu\relax
\input{abstract}
\endgroup

\input{main_body}

\bibliography{references}
\clearpage

\flushbottom
\appendix
\renewcommand{\theHfigure}{appendix.\arabic{figure}}
\renewcommand{\theHtable}{appendix.\arabic{table}}
\setcounter{figure}{0}
\renewcommand{\thefigure}{A\arabic{figure}}
\setcounter{table}{0}
\renewcommand{\thetable}{A\arabic{table}}
\twocolumn[{\centering\Large\bfseries Supplementary Material\par\vspace{1em}}]
\input{supplementary}

\end{document}

%% file: supplementary_style.tex
\definecolor{AcademicNavy}{HTML}{293B67}
\definecolor{AcademicBlueGray}{HTML}{64748B}
\definecolor{AcademicLavender}{HTML}{F2F3F9}
\definecolor{AcademicTeal}{HTML}{356B70}
\definecolor{AcademicMint}{HTML}{EFF6F5}
\definecolor{CaseBlue}{HTML}{3B5F82}
\definecolor{CaseTeal}{HTML}{3F746F}
\definecolor{CaseOchre}{HTML}{8A6935}
\definecolor{CasePlum}{HTML}{6C587B}
\definecolor{CaseNeutral}{HTML}{F6F7F9}
\definecolor{CaseRule}{HTML}{D4D9E1}
\definecolor{CodeText}{HTML}{000000}
\definecolor{CodeKeyword}{HTML}{4B168F}
\definecolor{CodeComment}{HTML}{005A20}
\definecolor{CodeString}{HTML}{004C9E}
\definecolor{CodeBackground}{HTML}{F8F8F8}
\definecolor{CodeRule}{HTML}{BEC7D2}
\definecolor{CodeNumber}{HTML}{424A55}

\newcolumntype{L}[1]{>{\RaggedRight\arraybackslash}p{#1}}
\newcolumntype{C}[1]{>{\Centering\arraybackslash}p{#1}}
\newcolumntype{Y}{>{\RaggedRight\arraybackslash}X}
\newcommand{\suppTableSetup}{%
  \small
  \setlength{\tabcolsep}{3.8pt}%
  \renewcommand{\arraystretch}{1.12}%
}
\newcommand{\suppDenseTableSetup}{%
  \small
  \setlength{\tabcolsep}{3.2pt}%
  \renewcommand{\arraystretch}{1.14}%
}
\newcounter{suppalgorithm}[section]
\renewcommand{\thesuppalgorithm}{\thesection.\arabic{suppalgorithm}}
\newenvironment{suppalgorithm}[2][]{%
  \refstepcounter{suppalgorithm}%
  \begin{tcolorbox}[
    enhanced,
    breakable,
    colback=AcademicLavender!58!white,
    colframe=AcademicNavy!72,
    colbacktitle=AcademicNavy!11,
    coltitle=black,
    fonttitle=\bfseries\small,
    title={Algorithm \thesuppalgorithm: #2},
    boxrule=0.45pt,
    arc=1.2pt,
    left=6pt,
    right=6pt,
    top=5pt,
    bottom=5pt,
    before skip=8pt,
    after skip=8pt,
    #1
  ]%
  \small
}{%
  \end{tcolorbox}
}
\newenvironment{promptbox}[1]{%
  \begin{tcolorbox}[
    enhanced,
    breakable,
    colback=AcademicLavender,
    colframe=AcademicNavy,
    colbacktitle=AcademicNavy!12,
    coltitle=black,
    fonttitle=\bfseries\small,
    title={#1},
    boxrule=0.45pt,
    arc=1.2pt,
    left=7pt,
    right=7pt,
    top=6pt,
    bottom=6pt,
    before skip=8pt,
    after skip=8pt
  ]%
  \small
}{%
  \end{tcolorbox}
}
\newenvironment{contractbox}[1]{%
  \begin{tcolorbox}[
    enhanced,
    breakable,
    colback=AcademicLavender!78!white,
    colframe=AcademicNavy!88,
    colbacktitle=AcademicNavy!12,
    coltitle=black,
    fonttitle=\bfseries\small,
    title={#1},
    boxrule=0.50pt,
    arc=1.2pt,
    left=7pt,
    right=7pt,
    top=6pt,
    bottom=6pt,
    before skip=8pt,
    after skip=8pt
  ]%
  \small
}{%
  \end{tcolorbox}
}
\newenvironment{readingbox}[1]{%
  \begin{tcolorbox}[
    enhanced,
    breakable,
    colback=AcademicMint,
    colframe=AcademicTeal!88,
    colbacktitle=AcademicTeal!13,
    coltitle=black,
    fonttitle=\bfseries\small,
    title={#1},
    boxrule=0.50pt,
    arc=1.2pt,
    left=7pt,
    right=7pt,
    top=6pt,
    bottom=6pt,
    before skip=8pt,
    after skip=8pt
  ]%
  \small
}{%
  \end{tcolorbox}
}

\lstdefinestyle{LongSearchCode}{
  language=Python,
  basicstyle=\rmfamily\bfseries\fontsize{8.10}{8.60}\selectfont\color{CodeText},
  keywordstyle=\color{CodeKeyword}\bfseries,
  commentstyle=\color{CodeComment},
  stringstyle=\color{CodeString},
  identifierstyle=\color{CodeText},
  numbers=left,
  numberstyle=\rmfamily\bfseries\fontsize{7.00}{8.60}\selectfont\color{CodeNumber},
  numbersep=6pt,
  xleftmargin=1.55em,
  framexleftmargin=1.25em,
  breaklines=true,
  breakatwhitespace=false,
  columns=fullflexible,
  keepspaces=true,
  showstringspaces=false,
  literate={-}{{{\rmfamily\bfseries\selectfont -}}}1,
  tabsize=4,
  aboveskip=0pt,
  belowskip=0pt
}
\newtcblisting{codepanel}[2][]{%
  enhanced,
  listing only,
  listing engine=listings,
  listing options={style=LongSearchCode},
  colback=CodeBackground,
  colframe=CodeRule,
  colbacktitle=AcademicNavy!10!white,
  coltitle=black,
  fonttitle=\bfseries\footnotesize,
  title={#2},
  boxrule=0.48pt,
  arc=1.2pt,
  left=1pt,
  right=1pt,
  top=1.5pt,
  bottom=1.5pt,
  before skip=4pt,
  after skip=5pt,
  #1
}

\newcommand{\suppEntry}[2]{%
  \Needspace{3\baselineskip}%
  \par\smallskip
  \noindent\textbf{#1.} #2\par
}

\newcommand{\methodname}{\textsc{LongSearchEval}}
\newcommand{\methodabbr}{\textsc{LongSearchEval}}

%% file: abstract.tex
\begin{abstract}

Long-horizon web search requires LLM-based agents to integrate information across webpages through extended interactions under task-specific constraints. Evaluating this capability is important not only for measuring final answer quality, but also for assessing whether an agent acquires, verifies, and uses evidence reliably over extended interactions. Existing evaluations, however, mainly rely on simulated trajectories or on post-hoc trajectories reconstructed from final answers, rather than on retained real interaction traces, which limits the fidelity of behavioral assessment and weakens support for fine-grained error localization. To address this gap, we introduce \textsc{LongSearchEval}, a trajectory-aware agent evaluation framework that combines a unified agent execution protocol with source-constrained outcome verification and retrospective failure-node localization. Stage~1 verifies answer groups against agent-cited webpages without human-authored gold answers; Stage~2 audits normalized rounds and links flagged failure nodes to incorrect answer groups. We evaluate 16 models from eight families on 100 tasks, yielding 1,600 retained runs and 6,545 automatically flagged failure nodes. Trajectory collection and evaluation required 1,037 and 318 hours of cumulative runtime, respectively. Human validation yields 94.50\% agreement on supplied answer-group labels and 98.11\% feature-correspondence precision for audited round--group links. Together, outcome scores and linked trajectory evidence support controlled model comparison and retrospective inspection of answer-related content.

\end{abstract}

%% file: main_body.tex
\begin{figure*}[t]
\centering
\PaperFigure{0.95\textwidth}{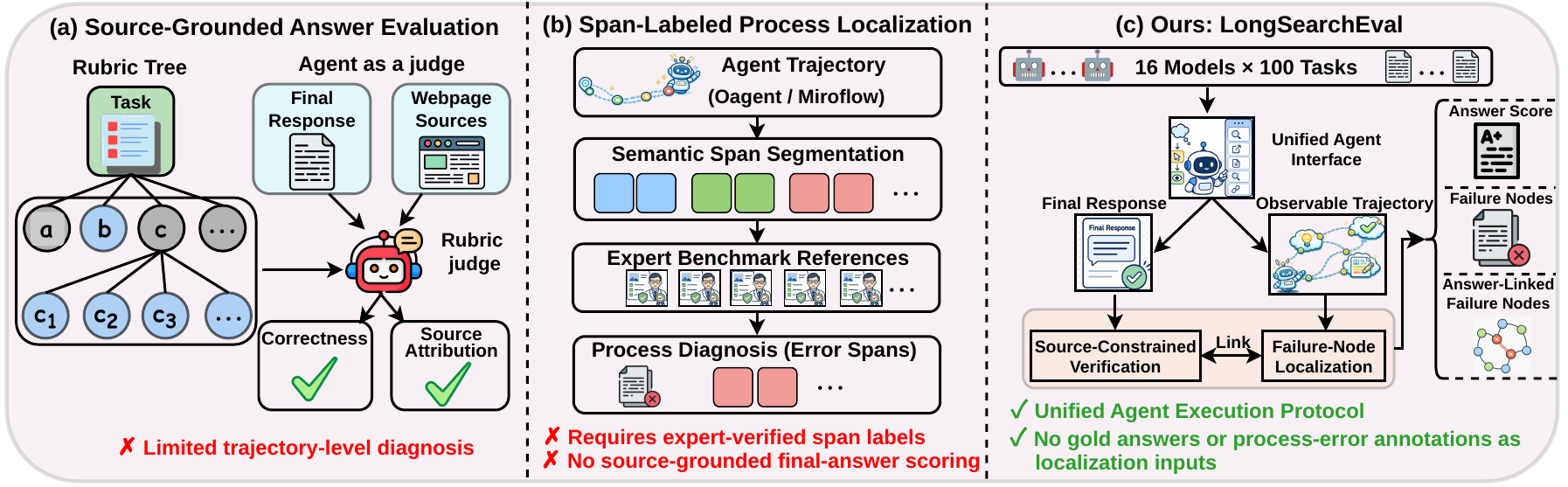}{Conceptual comparison}
\caption{
Source-grounded answer evaluation, process benchmarks, and retrospective
trajectory inspection. Expert labels in (b) are benchmark references, not
required localization inputs; (c) returns content related to Stage~1-incorrect
answer groups under a shared unified agent execution protocol.
}
\label{fig:intro_comparison}
\end{figure*}

\section{Introduction}

Large language models (LLMs) have demonstrated remarkable capabilities in
language understanding, generation, and complex reasoning, enabling a broad
range of applications beyond traditional natural language processing\cite{brown2020language}. Building on these capabilities, LLM-based agents extend models from passive
generation to autonomous problem solving by interacting
with external environments
\cite{yao2023react,wang2024mint}. Web search enables language models to acquire external and up-to-date
information by retrieving and inspecting web sources beyond their parametric
knowledge
\cite{nakano2021webgpt}. Agentic web search further transforms retrieval into autonomous information
seeking, where agents decompose complex queries, navigate webpages, verify
evidence, and synthesize citation-supported responses
\cite{gou2025mind2web2}.

Early approaches have explored complementary directions toward autonomous web
information seeking. Mind2Web~\cite{deng2023mind2web}, WebArena~\cite{zhou2024webarena},
and AgentBench~\cite{liu2024agentbench} establish general benchmarks for
real-world websites, reproducible browser environments, and multi-domain agent
evaluation. WebVoyager~\cite{he2024webvoyager}, WebLINX~\cite{lu2024weblinx},
and VisualWebArena~\cite{koh2024visualwebarena} extend evaluation to
multimodal or dialogue-conditioned web interaction. AssistantBench~\cite{yoran2024assistantbench},
WorkArena~\cite{drouin2024workarena}, WorkArena++~\cite{boisvert2024workarenaplus}, and TravelPlanner~\cite{xie2024travelplanner} examine realistic, time-consuming,
or compositional tasks that require extended information gathering and
planning. AppWorld~\cite{trivedi2024appworld}, OSWorld~\cite{xie2024osworld},
AndroidWorld~\cite{rawles2025androidworld}, and $\tau$-bench~\cite{yao2025taubench} further introduce stateful applications,
open-ended computer environments, and tool--user interaction
protocols. Collectively, these benchmarks substantially improve the
realism and scope of agent evaluation, while leaving limited support for
jointly verifying source-grounded final answers and locating related evidence
in comparable open-web trajectories.

Recent research has increasingly focused on evaluating \textsc{Deep\-Search}
agents from two complementary perspectives. The first direction emphasizes
source-grounded final-answer evaluation (\textbf{Figure~1(a)}). Mind2Web
2~\cite{gou2025mind2web2} introduces realistic long-horizon agentic-search
tasks and rubric-based Agent-as-a-Judge evaluation. GAIA~\cite{mialon2024gaia},
AssistantBench~\cite{yoran2024assistantbench}, LiveResearchBench~\cite{wang2025liveresearchbench}, and BrowseComp~\cite{wei2025browsecomp} evaluate complex information
seeking under external-knowledge, live-web, or persistent-browsing
requirements. RAGChecker~\cite{ru2024ragchecker}, long-form factuality~\cite{wei2024longformfactuality}
evaluation, and RAGTruth~\cite{niu2024ragtruth} provide finer-grained analyses of retrieval support,
claim correctness, and hallucination in grounded generation.
Because these settings often rely on LLM-based evaluators, a parallel line of
work studies judge construction~\cite{zheng2023judging} and meta-evaluation~\cite{zeng2024llmbar} through pairwise
comparison, adversarial instruction-following cases, critique generation~\cite{ke2024critiquellm},
rubric-conditioned scoring~\cite{kim2024prometheus}, hierarchical criteria~\cite{liu2024hdeval}, and judge-specific benchmarks~\cite{tan2025judgebench}. Other studies identify
systematic evaluator biases, including unfair comparison behavior~\cite{wang2024unfairevaluators} and
self-preference~\cite{panickssery2024selfpreference}. These methods substantially improve outcome
evaluation, but their primary object remains the generated answer and its
supporting evidence rather than the observable search process that produced
it. The second direction investigates process-level evaluation and error
localization (\textbf{Figure~1(b)}). Let's Verify Step by Step~\cite{lightman2024verify},
Math-Shepherd~\cite{wang2024mathshepherd}, and ProcessBench~\cite{zheng2025processbench} show that intermediate-step verification~\cite{lightman2024verify} can
reveal reasoning failures beyond final-answer accuracy.
AgentBoard~\cite{ma2024agentboard} introduces progress-level diagnostics for multi-turn agents, while
analyses of error snowballing~\cite{zhang2024snowballing} show how early mistakes can induce downstream
failures. TELBench and DRIFT~\cite{wang2026telbench} further
adapt process localization to DeepResearch trajectories by segmenting agent
trajectories and identifying harmful error spans. These approaches demonstrate the value of process-level diagnosis.
Human-verified labels support their evaluation benchmarks, which is distinct
from requiring such labels as localization inputs. Our focus is the outcome-conditioned connection
between answer groups judged incorrect and retained web-search content under
a unified agent execution protocol.

In response to these challenges, we introduce \textsc{LongSearchEval}, a trajectory-aware agent evaluation framework for assessing the long-horizon web-search capabilities of LLMs (\textbf{Figure~1(c)}). Each evaluated LLM operates within a shared agent controller and standardized web-tool interface, and its final response is assessed together with the observable trajectory that produced it. The shared controller and restricted tool outputs standardize the interaction conditions, connecting model-level comparison with inspectable evidence of search behavior.

\textsc{LongSearchEval} combines controlled trajectory collection with two evaluation stages. Source-constrained outcome verification checks final-answer groups using only the webpages cited by the agent, without requiring human-authored gold answers. Conditioned on the final response and Stage~1 judgments, retrospective failure-node localization links related content, sources, or fields to groups judged incorrect. The returned associations expose where
constraint-relevant evidence for each incorrect group appears in the trajectory. A correct webpage observation can still expose an answer discrepancy. These associations support retrospective inspection, not identification of an agent's first erroneous commitment.

We apply \textsc{LongSearchEval} to 16 models from eight families on 100 real-world web-search tasks, yielding 1,600 retained runs. Trajectory collection and evaluation required 1,037 and 318 hours, respectively. Human review of 946 supplied answer groups from 200 responses yields 94.50\% agreement with automatic judgments, while Answer Score and Answer Precision correlate with human scores at Pearson coefficients of 0.9054 and 0.9132. In a separate run sample, review of all 317 retained round--group links confirms 311 feature correspondences (98.11\%). The experiments also reveal distinct search and reading patterns among models with similar answer scores.

\noindent\textbf{Our contributions are threefold:}

\begingroup
\setlength{\leftmargini}{1.7em}
\renewcommand{\labelenumi}{\textbf{\arabic{enumi}.}}
\begin{enumerate}
\setlength{\topsep}{0.20em}
\setlength{\itemsep}{0.20em}
\setlength{\parsep}{0pt}
\setlength{\parskip}{0pt}

\item We formulate long-horizon web-search evaluation as a joint analysis of source-supported outcomes and observable trajectories under a controlled agent protocol.

\item We introduce \textsc{LongSearchEval}, which combines trajectory collection under a unified execution protocol, source-constrained verification, and retrospective localization of content related to Stage~1 incorrect groups, without gold answers or pre-labeled process errors as operational inputs.

\item We build a 1,600-run corpus covering 16 LLMs and 100 tasks, with human validation of 946 supplied answer groups and all 317 retained links in fixed samples.

\end{enumerate}
\endgroup

\section{Related Work}
\label{sec:related_work}

\textbf{Web Agent Evaluation.}
Web-agent evaluation has progressed from isolated browsing actions toward
realistic long-horizon interaction and evidence-grounded task completion.
Mind2Web~\cite{deng2023mind2web} introduces large-scale real-world website
tasks for generalist web agents, while WebArena~\cite{zhou2024webarena}
provides realistic multi-website environments for autonomous interaction.
AgentBench~\cite{liu2024agentbench} evaluates LLMs as agents across diverse
environments. WebVoyager~\cite{he2024webvoyager}, WebLINX
~\cite{lu2024weblinx}, and VisualWebArena~\cite{koh2024visualwebarena}
extend evaluation to multimodal or dialogue-conditioned web interaction.
AssistantBench~\cite{yoran2024assistantbench}, WorkArena
~\cite{drouin2024workarena}, WorkArena++~\cite{boisvert2024workarenaplus},
and TravelPlanner~\cite{xie2024travelplanner} further study realistic,
time-consuming, or compositional information-seeking tasks, while AppWorld
~\cite{trivedi2024appworld}, OSWorld~\cite{xie2024osworld},
AndroidWorld~\cite{rawles2025androidworld}, and $\tau$-bench
~\cite{yao2025taubench} investigate stateful environments and tool-user
interaction. Beyond task completion, recent studies have focused on evaluating whether agent
responses are supported by external evidence. Mind2Web 2~\cite{gou2025mind2web2}
introduces rubric-based Agent-as-a-Judge evaluation for citation-supported web
responses. GAIA~\cite{mialon2024gaia}, AssistantBench~\cite{yoran2024assistantbench},
LiveResearchBench~\cite{wang2025liveresearchbench}, and BrowseComp
~\cite{wei2025browsecomp} evaluate complex information seeking under
external-knowledge and live-web settings. RAGChecker~\cite{ru2024ragchecker},
long-form factuality evaluation~\cite{wei2024longformfactuality}, and RAGTruth
~\cite{niu2024ragtruth} further analyze retrieval support, claim correctness,
and hallucination. These methods improve outcome evaluation, but mainly focus
on final responses rather than indexed trajectory content related to
answer groups judged incorrect.

\textbf{Process-Level Diagnosis.}
Process-level evaluation aims to identify failures beyond final outputs.
Let's Verify Step by Step~\cite{lightman2024verify}, Math-Shepherd
~\cite{wang2024mathshepherd}, and ProcessBench~\cite{zheng2025processbench}
show that intermediate-step verification can reveal reasoning failures beyond
final-answer accuracy. AgentBoard~\cite{ma2024agentboard} introduces
progress-level diagnostics for multi-turn agents, while error snowballing
analysis~\cite{zhang2024snowballing} studies how early mistakes induce
downstream failures. TELBench and DRIFT~\cite{wang2026telbench} further extend
process localization to DeepResearch trajectories by segmenting agent
trajectories and identifying harmful error spans. These approaches demonstrate the value of process diagnosis.
\textsc{LongSearchEval} connects source-constrained answer-group verification
with retrospective localization of related observations, sources, and fields
under a unified agent execution protocol. The final response and its judgments
are inputs to this task; returned associations do not identify error onset.

\begin{figure*}[t]
\centering
\PaperFigure{0.90\textwidth}{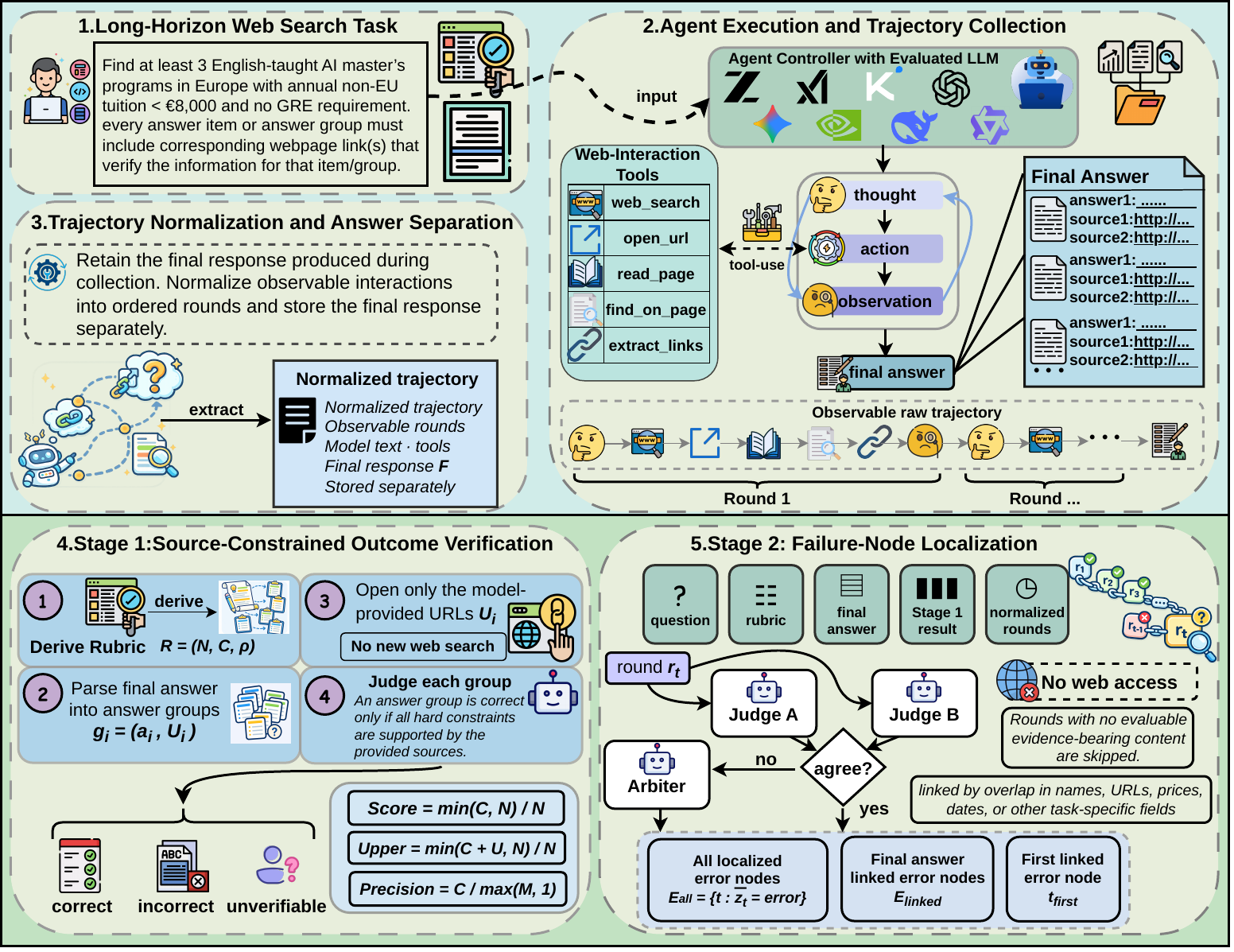}{Method overview}
\caption{
Overview of \textsc{LongSearchEval}. A unified agent interface records
each LLM's observable trajectory and final response. Stage~1 verifies answers
using agent-provided URLs. Conditioned on the response and Stage~1 judgments,
Stage~2 links related content in normalized rounds to incorrect answer groups
without web access or hidden chain-of-thought.
}
\label{fig:method_overview}
\end{figure*}

\section{Method}
\label{sec:method}

\textsc{LongSearchEval} evaluates LLMs' long-horizon web-search capabilities
through a shared agent execution protocol and two complementary evaluation
stages. For each model--task run, we jointly consider the source-grounded final
response \(F\) and the observable web-interaction trajectory \(\tau\) produced
while solving the task. Agent execution under a unified protocol first provides
comparable interaction records. Stage~1 then verifies source-supported answer
groups in \(F\), and Stage~2 uses these judgments and \(F\) to flag candidate
locations and link their content to groups judged incorrect by Stage~1.
\textbf{Figure~\ref{fig:method_overview}} illustrates the complete workflow.

\subsection{Agent Execution and Trajectory Collection}
\label{sec}

Within \textsc{LongSearchEval}, we run 16 LLMs on a shared task set and collect the complete web-search trajectories produced during actual agent execution. A representative example of the trajectory-collection workflow is illustrated in Appendix~\ref{app:trajectory_collection_case}.

Given model $m$ and task question $q$, we initialize the agent history as $\mathcal{H}_{m,q,0}=(q)$. At round $t$, the model generates an explicitly recorded reasoning state $h_{m,q,t}$ from the existing interaction history and then selects a tool action $a_{m,q,t}$. This sequential dependence is formalized as
\begin{equation}
\begin{aligned}
&p_{\theta_m}
\left(
h_t,a_t
\mid
\mathcal{H}_{t-1}
\right)
\\
&\quad=
p_{\theta_m}
\left(
h_t
\mid
\mathcal{H}_{t-1}
\right)
p_{\theta_m}
\left(
a_t
\mid
\mathcal{H}_{t-1},h_t
\right).
\end{aligned}
\label{eq:react_step}
\end{equation}

Here, $\theta_m$ denotes the parameters of model $m$. The variable $h_{m,q,t}$ corresponds to the observable textual reasoning field logged by the shared agent controller; it does not include unobserved internal states. A tool action is represented as $a_{m,q,t}=(w_{m,q,t},\xi_{m,q,t})$, where $w_{m,q,t}$ identifies a tool in the shared web-tool interface $\mathcal{W}$ and $\xi_{m,q,t}$ contains its invocation arguments. After the action is executed, the web environment returns an observation according to
\begin{equation}
o_{m,q,t}
\sim
p_{\mathcal{W}}
\left(
o
\mid
a_{m,q,t},
s_{m,q,t-1}^{\mathrm{web}}
\right),
\end{equation}
where $s_{m,q,t-1}^{\mathrm{web}}$ denotes the web-environment state before the action. Search results, webpage content, and tool failures are all recorded as observations $o_{m,q,t}$ in their original execution order. Web content therefore serves as environmental feedback within the agent interaction rather than as a separate collection objective. After round $t$, the interaction history is updated as
\begin{equation}
\mathcal{H}_{m,q,t}
=
\left(
q,
\left(
h_{m,q,s},
a_{m,q,s},
o_{m,q,s}
\right)_{s=1}^{t}
\right).
\end{equation}
Subsequent decisions use this history, which retains unsuccessful
searches, intermediate revisions, and candidates omitted from the
final response. Let $d_{m,q,t}$ denote the controller's termination
state after round $t$, and let $B$ be the maximum interaction budget.
The realized stopping round is
\begin{equation}
T_{m,q}
=
\min
\left\{
t\geq 1:
d_{m,q,t}=\mathrm{finalize}
\lor
t=B
\right\}.
\end{equation}
After round $T_{m,q}$, the agent generates the terminal response $F_{m,q}$ conditioned on the complete interaction history. The resulting raw trajectory is
\begin{equation}
\tau_{m,q}^{\mathrm{raw}}
=
\left(
\left(
h_{m,q,t},
a_{m,q,t},
o_{m,q,t}
\right)_{t=1}^{T_{m,q}},
F_{m,q}
\right).
\end{equation}

Conditioned on the task and realized trajectory length, the complete generation process factorizes as
\begin{equation}
\begin{aligned}
&p_{\theta_m,\mathcal{W}}
\left(
\tau^{\mathrm{raw}}
\mid
q,T
\right)
\\
&\quad=
p_{\theta_m}
\left(
F
\mid
\mathcal{H}_{T}
\right)
\prod_{t=1}^{T}
p_{\theta_m}
\left(
h_t,a_t
\mid
\mathcal{H}_{t-1}
\right) \\
&\qquad p_{\mathcal{W}}
\left(
o_t
\mid
a_t,s_{t-1}^{\mathrm{web}}
\right).
\end{aligned}
\label{eq:trajectory_factorization}
\end{equation}

After collection, the resulting trajectory preserves the complete observable decision process from the task question to the final response, allowing subsequent evaluators to inspect where task-relevant defects are flagged and how these rounds relate to the retained final response.

\subsection{Source-Constrained Outcome Verification}
\label{sec:outcome_verification}

Stage~1 evaluates how well the final response fulfills the task
requirements by verifying its answer groups against the webpages
cited in that response. For each group, the verifier checks whether
the cited content supports all applicable requirements and then
aggregates the group-level judgments into outcome scores. It uses
only agent-provided URLs and performs no additional web search,
enabling source-constrained evaluation without human-authored
gold answers.

For each retained run $(m,q)$, Stage~1 independently instantiates a rubric
\[
R_{m,q}
=
\left(
N_{m,q},
\mathcal{C}_{m,q},
\rho_{m,q}
\right)
\]
from the task question under the same evaluator configuration.
Here, $N_{m,q}$ is the required minimum number of answer groups,
$\mathcal{C}_{m,q}$ is the set of hard task constraints, and
$\rho_{m,q}$ specifies the grouping rule. The evaluated model identity is not
included in this generation call. Stage~1 then parses the final response into:
\begin{equation}
\begin{gathered}
\mathcal{G}(F_{m,q})
=
\left\{
g_{m,q,i}
\right\}_{i=1}^{P_{m,q}},
\\
g_{m,q,i}
=
\left(
a_{m,q,i},
\mathcal{U}_{m,q,i}
\right),
\end{gathered}
\end{equation}
where $a_{m,q,i}$ is the answer content of group $i$ and
$\mathcal{U}_{m,q,i}$ is the set of URLs cited as its evidence. The minimum
count $N_{m,q}$ is used for task-level scoring, whereas each proposed group must
independently satisfy all applicable hard constraints. For readability, we
suppress the run indices $(m,q)$ in the remainder of this subsection.
The count-mode conventions used in the experiment are specified in Section~\ref{sec:experimental_setup} alongside the score aggregation rules.

For each answer group $g_i=(a_i,\mathcal{U}_i)$, the verifier fetches only the
URLs in $\mathcal{U}_i$ and denotes the resulting webpage evidence by $E_i$.
Let $\beta_i=1$ when at least one cited URL yields usable page content.
Let $\gamma_i=1$ when $\mathcal{U}_i$ is non-empty but all cited URLs fail
solely because of source-access uncertainty, such as a timeout, temporary
unavailability, network failure, or anti-bot blocking. Missing, malformed, fabricated, or irrelevant citations are not treated as access-only failures under this criterion.

Missing citations, unsupported claims, invalid attributions, and
constraint violations are treated as incorrect, whereas groups that cannot be
verified solely because of source-access uncertainty are kept separate as
unverifiable.

Let $\mathcal{C}_i\subseteq\mathcal{C}$ contain the blocking constraints
applicable to group $g_i$, excluding the task-level count requirement. The
verifier defines a binary support function $\psi(g_i,c,E_i)$, where
$\psi(g_i,c,E_i)=1$ means that the cited evidence supports constraint $c$ for
group $g_i$. The group verdict is
\begin{equation}
v_i =
\begin{cases}
\mathrm{correct}, &
\begin{aligned}
&\beta_i=1
\\[-2pt]
&\;\wedge\;
\displaystyle\prod_{c\in\mathcal{C}_i}
\psi(g_i,c,E_i)=1,
\end{aligned}
\\[3pt]
\mathrm{unverifiable}, &
\beta_i=0
\;\wedge\;
\gamma_i=1,
\\[3pt]
\mathrm{incorrect}, &
\text{otherwise}.
\end{cases}
\end{equation}

Let
$C=\sum_i \mathbf{1}[v_i=\mathrm{correct}]$,
$U=\sum_i \mathbf{1}[v_i=\mathrm{unverifiable}]$, and
$M=|\mathcal{G}(F)|$. We define the normalized answer score
$s_{\mathrm{ans}}$, its optimistic counterpart $s_{\mathrm{ans}}^{+}$, and
answer precision $q_{\mathrm{ans}}$ as
\begin{equation}
\begin{aligned}
s_{\mathrm{ans}}     &= \tfrac{\min(C,N)}{N}, \\
s_{\mathrm{ans}}^{+} &= \tfrac{\min(C+U,N)}{N}, \\
q_{\mathrm{ans}}     &= \tfrac{C}{\max(M,1)} .
\end{aligned}
\end{equation}
The reported Answer Score and Upper Score are percentage scores,
$100s_{\mathrm{ans}}$ and $100s_{\mathrm{ans}}^{+}$, respectively. The main
answer score measures task-level requirement satisfaction under the
minimum-count setting; Upper Score treats the groups currently marked
unverifiable as potentially correct, and precision measures the verified fraction of the returned groups. These quantities measure satisfaction under
agent-provided evidence.

\subsection{Trajectory-Aware Failure-Node Localization}
\label{sec:trajectory_localization}

Stage~2 traces answer groups judged incorrect by Stage~1 back to
related webpage content in the recorded trajectory. For each such
group, the localization task is to identify relevant observations,
sources, or fields along the original round order, from the
beginning of the trajectory onward. The resulting round--group
associations provide concrete evidence locations for inspecting
the final answer, without re-evaluating its Stage~1 verdict.

Each judge receives only the task question, run-specific rubric, final response, Stage~1 answer-group verdicts, and one normalized round, with no web access or evaluated-model identity. Judges assess only observable content, not latent reasoning or hidden chain-of-thought. Each round is represented as
\begin{equation}
x_{m,q,t}
=
\left(
q,
R_{m,q},
F_{m,q},
V_{m,q}^{(1)},
r_{m,q,t}
\right),
\end{equation}
where
$
V_{m,q}^{(1)}
=
\left\{
\left(
g_{m,q,i},
v_{m,q,i}
\right)
\right\}_{i=1}^{P_{m,q}}
$
contains the Stage~1 verdicts; $(m,q)$ is omitted when unambiguous. Judges independently produce
\[
\hat{y}_t^j
=
\left(
{z}_t^j,
{\mathcal{L}}_t^j
\right),
\quad
j\in\{A,B\},
\]
where
${z}_t^j\in\allowbreak
\{\mathrm{error},\mathrm{nonerror},\mathrm{skipped}\}$
is the verdict and
$\hat{\mathcal{L}}_t^j$
the proposed answer-group links. For a jointly identified error,
$
\hat{\mathcal{L}}_t^{AB}
=
\left(
{\mathcal{L}}_t^A
\cup
{\mathcal{L}}_t^B
\right).
$
The consensus output
$\bar{y}_t=(\bar{z}_t,\bar{\mathcal{L}}_t)$
is
\begin{equation}
\bar{y}_t =
\begin{cases}
\bigl(
\mathrm{error},
\hat{\mathcal{L}}_t^{AB}
\bigr),
&
\hat{z}_t^A=\hat{z}_t^B=\mathrm{error},
\\[3pt]
\hat{y}_t^A,
&
\hat{z}_t^A=\hat{z}_t^B\neq\mathrm{error},
\\[3pt]
\left(
x_t,
\hat{y}_t^A,
\hat{y}_t^B
\right),
&
\hat{z}_t^A\neq\hat{z}_t^B.
\end{cases}
\end{equation}

Agreement does not trigger arbitration. Joint errors retain the union of proposed links and use Judge~A's normalized error type when available, otherwise Judge~B's. For differing verdicts, the third-case tuple is passed to the arbiter and its output is adopted. Rounds assigned \emph{error} by the arbiter remain flagged failure nodes when no valid normalized type is available, but are excluded from taxonomy-specific analyses.

The original candidate checks and \emph{error}/\emph{nonerror}/\emph{skipped} labels are retained, including checks on exploratory candidates. Extracted units and reasons are evaluator interpretations, not agent quotations. Inspecting an ineligible candidate is not itself erroneous; correct observations can expose answer discrepancies. A consensus-\emph{error} round is termed a \emph{failure node}, used as an automatic candidate flag. All such nodes form
$
\mathcal{E}_{\mathrm{all}}
=
\left\{
t:
\bar{z}_t=\mathrm{error}
\right\}.
$
We report $\mathcal E_{\mathrm{all}}$ as an automatic diagnostic inventory, not a behavioral-error rate; evaluable-round counts and organization depend on search behavior within each run.

Judges associate flagged failure nodes with final-answer groups through task-relevant semantic cues without requiring exact lexical overlap. The proposed answer-group links are standardized in format without
an additional deterministic semantic-consistency check, to accommodate
implicit correspondences in open-ended tasks. When both judges label
the same round as \emph{error}, their proposed links are merged and
deduplicated, while arbitration-resolved rounds retain the arbiter's links. Because raw diagnostic links may include groups not confirmed as incorrect by Stage~1, downstream analyses retain only links to the confirmed incorrect set
$
\mathcal{B}
=
\left\{
i\in\{1,\ldots,P\}:
v_i=\mathrm{incorrect}
\right\}
$.
Correct, unverifiable, and nonexistent targets are excluded. This target-status filter is not a semantic check; Stage~1 mistakes can affect the retained associations. Answer-linked failure nodes are
\begin{equation}
\mathcal{E}_{\mathrm{linked}}
=
\left\{
t\in\mathcal{E}_{\mathrm{all}}:
\bar{\mathcal{L}}_t
\cap
\mathcal{B}
\neq
\emptyset
\right\}.
\end{equation}
When $\mathcal{E}_{\mathrm{linked}}$ is non-empty, we define
$
t_{\mathrm{first}}
=
\min \mathcal{E}_{\mathrm{linked}}
$
as the earliest returned related location, not the first erroneous commitment or a causal origin. The nodes in $\mathcal{E}_{\mathrm{linked}}$ and their retained round--group edges are the main retrospective output; $\mathcal{E}_{\mathrm{all}}$ also includes unlinked flags. Nodes, edges, and target groups have separate denominators because links can be many-to-many. Appendix~\ref{app:answer_process_case} illustrates these source--answer comparisons.

Together, answer metrics and group verdicts characterize outcomes, while these three diagnostics summarize all flagged failure nodes, answer-linked failure nodes, and the earliest linked round. Auxiliary signals for search, content inspection, link navigation, and premature answering support complementary behavioral analyses. Both stages are model-identity blind; model IDs are attached only after judging.

\begin{table*}[t]
\caption{
Outcome quality, search behavior, evidence use, and execution robustness
evaluated with \textsc{LongSearchEval} under the controlled agent execution setting. AS, AP, and UGR denote Answer Score, Answer
Precision, and Unverifiable Group Rate. AP and UGR are percentages, and Read kChars is the mean character count
divided by 1,000. Valid and Prem. use the saved final-answer-validity and
premature-answer indicators. Models are grouped by family, with thin horizontal rules separating families to facilitate within- and cross-family comparisons.
}
\label{tab:main_results}
\centering
\begingroup
\small
\setlength{\tabcolsep}{2pt}
\renewcommand{\arraystretch}{1.00}

\begin{tabular*}{\textwidth}{
@{\extracolsep{\fill}}
l
cccc
ccc
cc
cc
@{}
}
\toprule
&
\multicolumn{4}{c}{\textbf{\shortstack{Outcome\\Quality}}} &
\multicolumn{3}{c}{\textbf{\shortstack{Search\\Behavior}}} &
\multicolumn{2}{c}{\textbf{\shortstack{Evidence\\Use}}} &
\multicolumn{2}{c}{\textbf{\shortstack{Execution\\Robustness}}}
\\
\cmidrule(lr){2-5}
\cmidrule(lr){6-8}
\cmidrule(lr){9-10}
\cmidrule(lr){11-12}

\textbf{Model} &
\textbf{\shortstack{AS\\$\uparrow$}} &
\textbf{\shortstack{AP\\(\%) $\uparrow$}} &
\textbf{\shortstack{Upper\\$\uparrow$}} &
\textbf{\shortstack{UGR\\(\%) $\downarrow$}} &
\textbf{\shortstack{Avg.\\Rounds}} &
\textbf{\shortstack{Avg.\\Searches}} &
\textbf{\shortstack{Avg.\\Domains}} &
\textbf{\shortstack{Read\\kChars}} &
\textbf{\shortstack{Final\\URLs}} &
\textbf{\shortstack{Valid\\(\%) $\uparrow$}} &
\textbf{\shortstack{Prem.\\(\%) $\downarrow$}}
\\
\midrule

GPT-6.1-Sol
& \textbf{68.80} & \textbf{68.34} & \textbf{75.35} & 5.32
& 17.74 & 9.49 & 6.59
& 26.79 & 10.44
& \textbf{100} & 0 \\

GPT-6-Luna-Pro
& \underline{65.88} & \underline{67.39} & \underline{73.08} & 6.12
& 18.41 & 14.57 & 5.38
& 16.58 & 8.11
& \textbf{100} & 0 \\

GPT-5.4
& 54.23 & 55.99 & 58.85 & 3.56
& 19.67 & 11.19 & 5.37
& 10.24 & \underline{12.90}
& \textbf{100} & 0 \\

GPT-5-Mini
& 47.17 & 47.17 & 51.13 & 4.90
& 25.07 & 16.35 & 5.08
& 4.78 & 11.24
& 92 & 5 \\

\midrule[0.3pt]

Gemini-3.1-Pro
& 38.51 & 38.76 & 50.15 & 10.99
& 24.57 & 16.64 & 5.72
& 24.64 & 7.99
& \textbf{100} & 0 \\

\makecell[l]{Gemini-2.5-Flash}
& 33.33 & 35.20 & 38.31 & 6.48
& 32.37 & 11.42 & 6.54
& 36.44 & 8.45
& 95 & 3 \\

\midrule[0.3pt]

Kimi-K2.6
& 51.86 & 51.84 & 58.51 & 7.05
& \underline{34.67} & \underline{21.98} & \textbf{9.16}
& 44.75 & 9.08
& \underline{99} & 0 \\

Kimi-K2.5
& 50.35 & 50.12 & 55.97 & 5.04
& 32.48 & 18.88 & \underline{8.47}
& \underline{44.80} & 9.21
& \textbf{100} & 0 \\

\midrule[0.3pt]

Qwen3.7-Plus
& 49.23 & 49.01 & 54.03 & 4.68
& 20.44 & 13.80 & 5.15
& 23.11 & 8.02
& 96 & 2 \\

Qwen3.6-Flash
& 37.32 & 37.90 & 45.20 & 8.55
& 20.34 & 15.90 & 4.73
& 21.42 & 7.53
& 98 & 0 \\

\midrule[0.3pt]

DeepSeek-V3.2
& 54.38 & 54.48 & 59.28 & 4.66
& 27.67 & 13.75 & 8.39
& 22.29 & 9.35
& \textbf{100} & 0 \\

DeepSeek-V4-Flash
& 53.90 & 53.78 & 58.69 & 5.43
& 30.22 & \textbf{21.99} & 6.75
& 37.60 & 12.26
& \underline{99} & 0 \\

\midrule[0.3pt]

GLM-5
& 47.54 & 46.44 & 53.67 & 5.39
& 30.04 & 13.58 & 7.69
& \textbf{48.02} & 8.18
& \underline{99} & 0 \\

GLM-4.7-Flash
& 32.87 & 32.61 & 39.57 & 6.07
& 31.80 & 20.85 & 6.18
& 10.27 & 7.43
& 97 & 0 \\

\midrule[0.3pt]

Grok-4.20
& 48.31 & 48.04 & 52.93 & 4.33
& 25.40 & 10.67 & 6.87
& 34.43 & \textbf{14.34}
& \underline{99} & 1 \\

\midrule[0.3pt]

\makecell[l]{Nemotron-3-Ultra}
& 43.35 & 43.14 & 48.53 & 5.11
& \textbf{35.69} & 13.76 & 7.07
& 39.58 & 9.42
& \underline{99} & 0 \\

\bottomrule
\end{tabular*}
\endgroup

\end{table*}

\section{Experiments}
\label{sec:experiments}

We use \textsc{LongSearchEval} to compare LLMs' long-horizon web-search
performance through outcome scores, observable search behavior, and
failure-node diagnostics, then assess outcome verification and diagnostic
correspondence through human review.

\subsection{Experimental Setup}
\label{sec:experimental_setup}

\textbf{Tasks, models, and protocol.} The evaluation comprises 16 models from eight families on 100 long-horizon
web-search tasks: 46 from Mind2Web~2\cite{gou2025mind2web2}, 20 from LiveResearchBench\cite{wang2025liveresearchbench},
29 from AssistantBench\cite{yoran2024assistantbench}, and 5 manually curated tasks.
All models use the same agent controller, prompt, five
web-interaction tools, configured iteration limit, and execution
policy. Each model--task pair contributes
one retained run, giving 1,600 trajectories; recovery attempts do not create
additional samples. The evaluated DeepSeek models are V3.2 and V4-Flash.
Stage~1 uses DeepSeek-V4-Pro as its single verifier. Stage~2 uses
DeepSeek-V4-Pro as Judge~A and Qwen3.7-Max as Judge~B and arbiter.
This role assignment is fixed across all evaluated models and tasks;
evaluated-model identities are omitted from judging inputs.
The ten-category vocabulary is fixed from a disjoint 50-task,
500-trajectory development set.

\begin{figure*}[t]
\centering
\PaperFigure{0.95\textwidth}{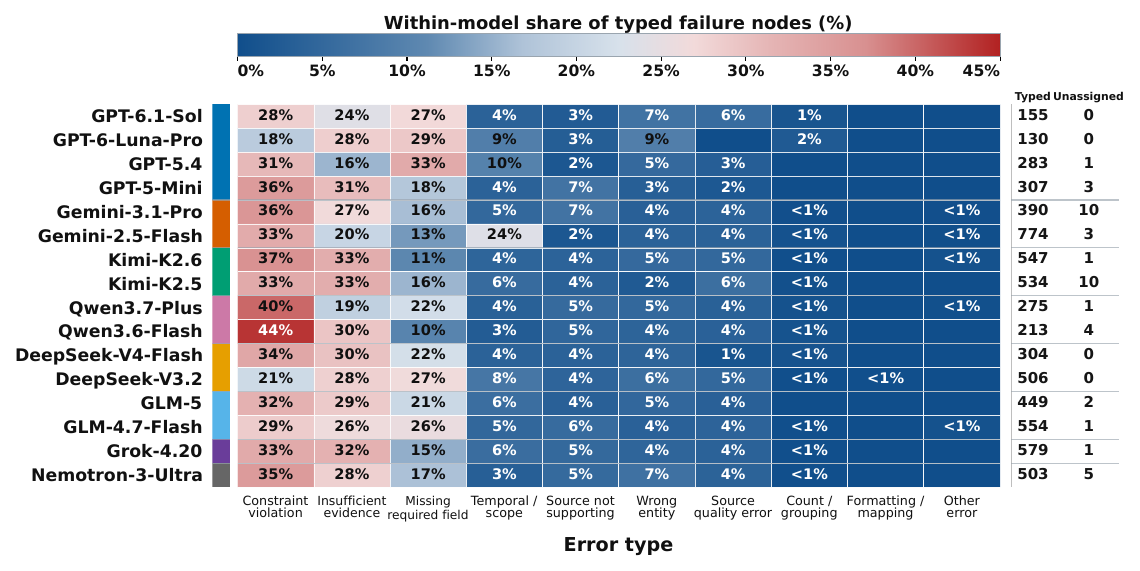}{Automatic error-type composition}
\caption{Diagnostic type composition of 6,503 automatically flagged and typed
candidate nodes. Each model is
normalized by its own typed-node count; 42 unassigned nodes remain in
localization totals but not in type denominators.}
\Description{Heatmap of ten error-type shares for each of the 16 models. Printed cell percentages accompany the color encoding. Counts at the right distinguish automatically typed and unassigned candidate nodes. Each row is normalized by the corresponding model's typed-node count.}
\label{fig:error_composition}
\end{figure*}

\textbf{Metrics and aggregation.} Answer Score (AS), Answer Precision (AP), and Upper Score are macro-averaged
over each model's runs; Unverifiable Group Rate pools $U/M$ over all
returned groups. The corpus contains 7,685 answer groups and 42,658
normalized rounds. Table~\ref{tab:main_results} jointly reports outcome,
behavior, evidence-use, and execution indicators.
The minimum-count formula in Section~\ref{sec:outcome_verification} applies
to \texttt{min}; the stored \texttt{exact} normalization uses $\max(N,M)$
and \texttt{dynamic\_all} uses $M$ as the AS/Upper denominator.
AP uses $\max(M,1)$, and confirmed empty outputs receive zero.
These conventions retain the exported scores without changing the labels.
Reported 95\% intervals use 10,000 paired task-bootstrap resamples with
fixed automatic outputs; they quantify task-level uncertainty rather than
repeated agent execution.

\subsection{Long-Horizon Web Search Performance}
\label{sec:overall_performance}

Table~\ref{tab:main_results} shows substantial variation under the shared
interface. GPT-6.1-Sol has the highest point estimates for AS (68.80),
AP (68.34\%), and Upper (75.35), followed by GPT-6-Luna-Pro in AS
(65.88) and AP (67.39\%). GPT-5.4 has the lowest UGR (3.56\%).
The Sol--Luna AS difference is 2.92 points, with a paired 95\% interval
of $[-5.67,11.99]$, whereas the GLM-5--GLM-4.7-Flash difference is
14.67 points $[8.41,20.82]$. Thus the results distinguish broad
performance gaps from closely spaced point estimates.

\begin{figure}[!t]
\centering
\includegraphics[width=1\linewidth]{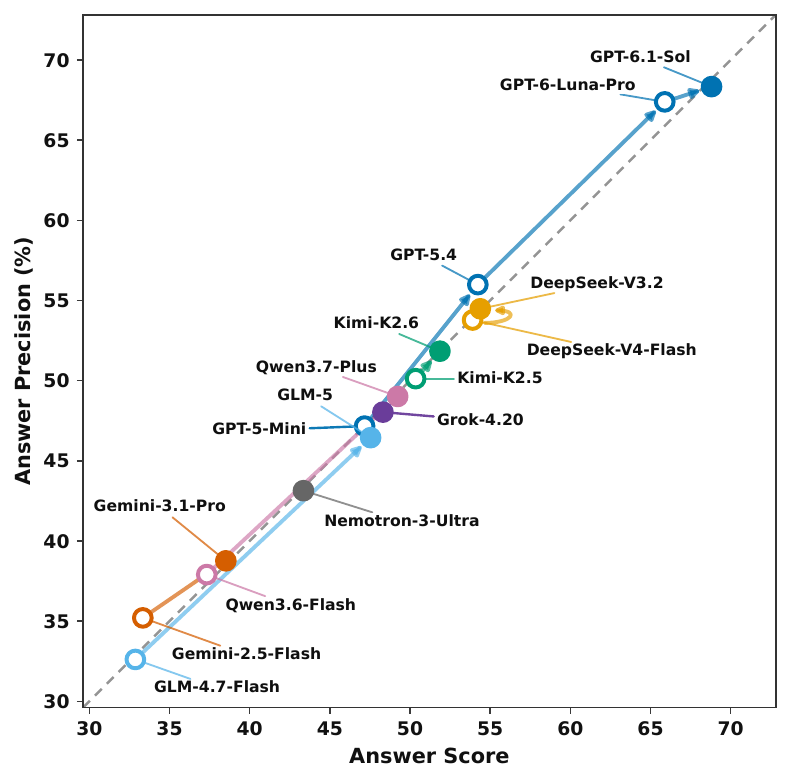}
\caption{Answer Score versus Answer Precision for 16 LLMs. Colors and lines denote
model families.}
\Description{Scatter plot of Answer Score against Answer Precision for the 16 evaluated models, with model names next to the points and within-family connecting lines. GPT-6.1-Sol and GPT-6-Luna-Pro appear toward the upper right. The axes and caption define the two reported outcome measures.}
\label{fig:score_precision}
\end{figure}

Similar outcome scores can conceal different search profiles.
DeepSeek-V3.2 and V4-Flash obtain AS values of 54.38 and 53.90,
respectively, but average 13.75 versus 21.99 searches and 22.29k versus
37.60k read characters---about 37.5\% fewer searches and 40.7\% less
text for V3.2. The paired AS difference is 0.48 points
$[-6.03,6.95]$. Across models, search counts range from 9.49 to 21.99,
and read text from 4.78k to 48.02k characters. Grok-4.20 returns the
most final URLs (14.34 per run) without leading the outcome scores.
Joint reporting therefore distinguishes verification success from
interaction volume and evidence presentation.

All 1,600 runs remain in the analysis, including 1,567 direct agent
responses, 20 collection-time no-tool syntheses, and 13 fallbacks.
The saved validity indicator is true for 1,573 responses (98.31\%);
11 runs carry a premature-answer flag. These retained paths are evaluated unchanged rather than limiting the comparison to successful executions with valid outputs.

\subsection{Failure-Node and Trajectory Diagnostics}
\label{sec:process_diagnostics}

Of 42,658 rounds, the final consensus retains 14,734 evaluable records:
6,545 automatic candidate flags and 8,189 \emph{nonerror} labels. Among flagged nodes, 3,664
(56.0\%) have retained links to Stage~1-incorrect answer groups.
The 5,798 distinct round--group edges cover 2,229 of 3,374 incorrect
groups, leaving 1,145 without a retained content association.
Table~\ref{tab:process_diagnostics} reports node-level diagnostics
separately from answer-group coverage.

\raggedbottom
\setlength{\textfloatsep}{12pt plus 2pt minus 2pt}
\begin{table}[!t]
\caption{Automatic candidate-node diagnostics under Stage~1-incorrect-group
filtering. Earliest-association percentages use the 850 runs with retained links.}
\label{tab:process_diagnostics}
\centering
\small
\setlength{\tabcolsep}{3pt}
\begin{tabularx}{\linewidth}{@{}Xr@{}}
\toprule
\textbf{Process diagnostic} & \textbf{Result} \\
\midrule
All automatically flagged candidate nodes & 6,545 \\
Linked / unlinked candidate nodes & 3,664 / 2,881 \\
Typed / unassigned candidate nodes & 6,503 / 42 \\
Unlinked incorrect final-answer groups & 1,145 \\
Runs with at least one linked candidate node & 850 \\
Earliest association by round 5 / 20 & 46.5\% / 88.7\% \\
\bottomrule
\end{tabularx}

\end{table}

Among the 850 linked runs, the earliest returned association is by round~5
in 395 runs (46.5\%) and by round~20 in 754 (88.7\%); the median is
round~6, with an interquartile range of 3--13. These locations identify
related content for retrospective inspection, not when the agent first
made an erroneous commitment. Constraint violation (2,124 nodes), insufficient evidence
(1,782), and missing required field (1,222) account for 78.86\% of typed
nodes. \textbf{Figure~\ref{fig:error_composition}} shows their model-specific
composition: the constraint-violation share is 21.15\% for V3.2 and
33.55\% for V4-Flash despite their similar aggregate AS.
The taxonomy summarizes the judges' diagnostic categories over
candidate content. Human validation below assesses correspondence between the returned locations and their target answer groups through a manual audit of all retained links in the sampled trajectories.

\begin{figure*}[t]
\centering
\includegraphics[width=0.90\textwidth]
  {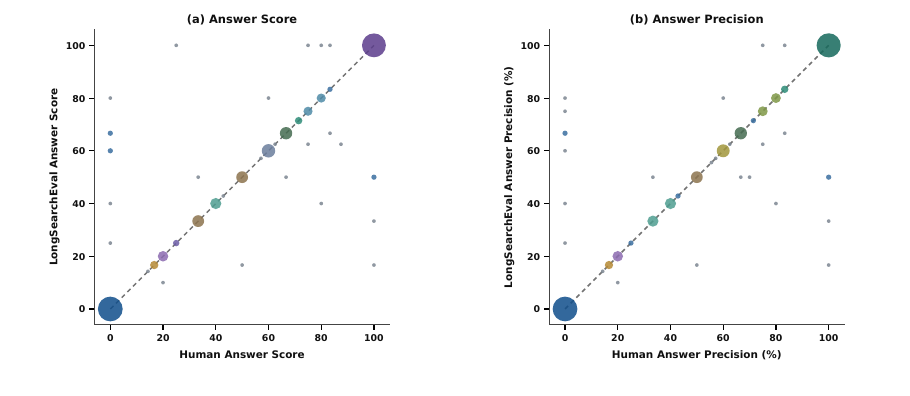}
\caption{
Automatic--human outcome-score alignment in the post-hoc
validation study: (a) Answer Score and (b) Answer Precision.
Both panels use the same 200 response pairs from 100 tasks.
Dashed lines indicate exact agreement ($y=x$);
marker area is proportional to the number of responses
sharing the same score pair.
}
\Description{Two bubble plots compare human scores on the horizontal axes with automatic LongSearchEval scores on the vertical axes. The left panel shows Answer Score and the right panel shows Answer Precision. Diagonal dashed lines mark exact agreement; larger bubbles represent more responses at the same score pair.}
\label{fig:human_score_alignment}
\end{figure*}

\subsection{Evaluator Reliability and Human Validation}
\label{sec:evaluator_reliability}

\textbf{Audit design.} The post-hoc audit covers \textbf{946 supplied answer groups across all tasks}, with two randomly sampled model responses per task.
Human labels serve as validation references, not inputs to scoring or
localization.
The primary reviewer (H1) checks requirements and the original cited URLs,
assigns correct/incorrect/unverifiable labels, and derives response-level
scores. Automatic verdicts, scores, reasons, and model metadata are
withheld until annotation is locked. The review retains the supplied
group boundaries, all 200 responses (including four confirmed empty-group
responses), and both sides' recorded quantity parameters.
A second reviewer (H2) independently checks a fixed 40-response, 172-group subset. Reported intervals use 10,000 paired task-cluster bootstrap resamples, retaining each task's two responses and all associated group-level records.

\textbf{Outcome agreement.}
Table~\ref{tab:human_correlations} reports Pearson $r$, Spearman $\rho$,
and Kendall $\tau$ on the same 200 response pairs. At group level,
894/946 judgments agree (94.50\%; $\kappa=0.8989$).
AS and AP have mean absolute errors of 4.88 score points and 4.45
percentage points, with machine-minus-human biases of 0.84 and 0.46,
respectively. The AS and AP Pearson 95\% intervals are
$[0.8496,0.9519]$ and $[0.8595,0.9588]$.
The two human reviewers agree on 160/172 subset group labels
(93.02\%; $\kappa=0.8574$).
\textbf{Figure~\ref{fig:human_score_alignment}} complements these statistics
by showing repeated exact matches across score levels and
lower-frequency off-diagonal discrepancies. The agreement in group judgments and response-level scores
supports the validity of source-constrained evaluation
on the supplied answer groups.

\begin{table}[htbp]
\caption{Score correlations following the 946-group human audit.
Each coefficient uses 200 response pairs from 100 tasks.}
\label{tab:human_correlations}
\centering
\small
\setlength{\tabcolsep}{4pt}
\renewcommand{\arraystretch}{1.10}
\begin{tabular}{@{}lccc@{}}
\toprule
\textbf{Score} & \textbf{Pearson $r$} & \textbf{Spearman $\rho$} & \textbf{Kendall $\tau$} \\
\midrule
Answer Score & 0.9054 & 0.9035 & 0.8609 \\
Answer Precision & 0.9132 & 0.9103 & 0.8712 \\
\bottomrule
\end{tabular}

\end{table}

\textbf{Within-task comparison.}
When both comparisons are non-tied, automatic and human orderings agree
on 56/60 tasks for AS and 57/61 for AP. Including 28/27 shared ties gives
84/100 matching relations for each metric; each also has four reversed
non-ties and 12 one-sided ties. These results assess within-task comparisons between the two
sampled model responses.

\textbf{Round-to-answer correspondence.}
The process sample covers all 2,642 normalized rounds of 100 runs from
50 randomly selected tasks. All 317 retained round--group links are reviewed,
including every retained edge in the sample. A supported link shares a
verified, locatable feature such as an entity, URL, provider, or price;
topic similarity alone is insufficient. For split/merged groups, at least
one mapped target must satisfy this rule. Review confirms 311/317 links,
giving \textbf{98.11\% feature-correspondence precision} (95\% CI:
$[96.15,99.40]$). The audit supports high-precision localization of answer-related
trajectory content under the defined feature-correspondence
criterion.

\textbf{Audit criteria.} A separate audit using an explicit erroneous-commitment criterion
assigned no error labels to the reviewed rounds, whereas the
candidate-content procedure produced 349 flags. These counts
concern different annotation targets. The returned-link audit
above evaluates the outcome-conditioned content-localization
task defined in Section~3.3.

\textbf{Internal consistency.}
Judges~A and~B agree on 12,417 of 14,818 round-level verdict
pairs (83.80\%). Among 4,612 rounds jointly labeled as
\emph{error} with comparable normalized types, their type
judgments agree in 4,032 cases (87.42\%). Together with human
validation of answer scores and returned-link correspondence,
these results provide empirical support for the effectiveness
and reliability of \textsc{LongSearchEval} in source-constrained
evaluation and retrospective trajectory-content localization.

\section{Conclusion}
\label{sec:conclusion}

We presented \textsc{LongSearchEval}, a trajectory-aware agent evaluation
framework for long-horizon web search that combines agent execution under a unified protocol, source-constrained outcome verification, and retrospective
failure-node localization. The 16-model, 1,600-run corpus combines answer scores, search behavior, and inspectable round-group associations. Similar answer scores can accompany different search workloads.
Conditioned on the final response and Stage~1 judgments,
localization connects answer groups with inspectable trajectory content.
Correct webpage observations can still be diagnostically useful when
compared with a group judged incorrect; the full candidate inventory is
reported separately from returned answer-linked evidence.
Human validation demonstrates close alignment with reference
answer judgments and high precision in returned content
correspondences. Together with cross-judge agreement, these
findings support the effectiveness and reliability of
\textsc{LongSearchEval} for source-constrained outcome evaluation
and retrospective trajectory inspection.

%% file: supplementary.tex
\input{supplementary_layout}
\setlength{\parskip}{1.8pt plus 0.6pt minus 0.2pt}
\input{supplementary_sections/01_scope}
\input{supplementary_sections/02_setup}
\input{supplementary_sections/03_collection}
\input{supplementary_sections/04_normalization}
\input{supplementary_sections/05_stage1}

\input{supplementary_sections/06_stage2}
\input{supplementary_sections/07_taxonomy}

\input{supplementary_sections/08_results}
\input{supplementary_sections/09_validation}
\input{supplementary_sections/08b_resources}

\input{supplementary_sections/11_reproduction}

\input{supplementary_sections/12_limitations}

%% file: supplementary_layout.tex
\captionsetup{
  hypcap=false,
  font=small,
  labelfont=bf,
  justification=justified,
  singlelinecheck=false,
  margin=0pt,
  skip=5pt
}
\setlength{\textfloatsep}{11pt plus 2pt minus 2pt}
\setlength{\floatsep}{10pt plus 2pt minus 2pt}
\setlength{\intextsep}{10pt plus 2pt minus 2pt}
\setlength{\dbltextfloatsep}{10pt plus 2pt minus 2pt}
\setlength{\dblfloatsep}{9pt plus 2pt minus 2pt}
\setlength{\emergencystretch}{1.5em}
\setlength{\jot}{4pt}
\renewcommand{\topfraction}{0.90}
\renewcommand{\dbltopfraction}{0.95}
\renewcommand{\dblfloatpagefraction}{0.80}
\renewcommand{\bottomfraction}{0.90}
\renewcommand{\textfraction}{0.05}
\renewcommand{\floatpagefraction}{0.80}
\flushbottom
\clubpenalty=10000
\widowpenalty=10000
\displaywidowpenalty=10000

\makeatletter
\setlength{\@fptop}{0pt}
\setlength{\@fpsep}{12pt plus 2pt}
\makeatother

\setlist[enumerate]{
  leftmargin=*,
  itemsep=1pt,
  parsep=0pt,
  topsep=3pt
}
\setlist[itemize]{
  leftmargin=*,
  itemsep=1pt,
  parsep=0pt,
  topsep=3pt
}

\urlstyle{rm}
\def\UrlFont{\rmfamily}
\DeclareUrlCommand\path{\urlstyle{rm}}
\Urlmuskip=0mu\relax

%% file: supplementary_sections/01_scope.tex
\section{Overview and Evaluation Units}
\label{app:supplement_guide}\label{app:supplement_crosswalk}
This supplement details the \methodname{} evaluation protocol and results. It follows
the same 16-model, eight-family, 100-task comparison, with one retained run per
model--task pair. It expands the collection protocol, source-constrained
verification, retrospective content localization, and validation. Human review
is a separate validation study, not an input to automatic evaluation.

\begin{readingbox}{How to interpret the localization output}
A \emph{failure node} is a normalized round assigned the stored consensus label
\emph{error}; it is an automatic candidate flag. A retained link associates that
round with an answer group judged incorrect by Stage~1. These associations
locate answer-related content in the recorded trajectory. Human review
separately evaluates returned feature correspondences and explicit agent
commitments.
\end{readingbox}

\subsection{Reading Guide}
Tasks and evaluator roles appear in Appendix~\ref{app:task_construction};
collection and normalization in Appendices~\ref{app:weak_tool_collection}
and~\ref{app:trajectory_construction}. Stage~1 and its rubric-consistency
analysis appear in Appendix~\ref{app:stage1_verification}; Stage~2, its
round accounting, and judge agreement appear in
Appendix~\ref{app:stage2_localization}.
Appendix~\ref{app:error_taxonomy} brings all ten error-type definitions,
four representative cases, and the type distribution together.
Appendix~\ref{app:extended_results} gives the complete quantitative diagnostics,
followed by human validation in Appendix~\ref{app:evaluator_reliability}
and experimental runtime and API budgets in Appendix~\ref{app:resource_costs}.
Appendix~\ref{app:reproduction} specifies prompt contracts and reproduction
procedures; Appendix~\ref{app:limitations} discusses evaluation scope.

\subsection{Objects and Denominators}
\label{app:evaluation_objects}\label{app:aggregation_conventions}
For model $m$ and task $q$, let $F_{m,q}$ be the retained final response and
$\tau_{m,q}=(r_1,\ldots,r_T)$ the normalized trajectory. The terminal response
is stored separately and never counted as a round. Each run has its own stored
rubric $R_{m,q}=(N,\mathcal C,\rho)$, specifying the required answer count,
hard constraints, and grouping rule, together with a count mode.
Stage~1 extracts groups $g_i=(a_i,\mathcal U_i)$, comprising answer content and
cited URLs, and labels each group correct, incorrect, or unverifiable.

\begin{table*}[tbp]
\suppTableSetup
\begin{tabularx}{\linewidth}{@{}L{0.23\linewidth}Y@{}}
\toprule
\textbf{Unit} & \textbf{Meaning and aggregation} \\
\midrule
Task / retained run & 100 task IDs; 1,600 model--task cells. Retries and resumes do not add samples. \\
Answer group & 7,685 supplied groups. AS, AP, and Upper are run-level scores macro-averaged over 100 tasks per model; UGR pools groups. \\
Normalized round & 42,658 ordered units; distinct from raw controller iterations and individual tool calls. \\
Flagged node & 6,545 consensus-error rounds, including 42 without an assigned type. Counts summarize the automatic diagnostic inventory. \\
Retained edge & A distinct run--round--incorrect-group triple. 5,798 edges connect 3,664 nodes to 2,229 groups. Each unit has its own denominator. \\
Earliest association & Minimum linked-round index, conditional on a retained link; defined for 850 runs. The other 750 receive no index. \\
\bottomrule
\end{tabularx}
\caption{Evaluation units and their separate aggregation denominators.}
\label{tab:evaluation_units}
\end{table*}
Type shares use typed nodes only; human-review denominators are specified
separately in Appendix~\ref{app:human_validation}.

%% file: supplementary_sections/02_setup.tex
\section{Tasks, Models, and Evaluator Configuration}
\label{app:task_construction}

The evaluation set targets long-horizon web search. Retained tasks
require sustained information seeking over public webpages, explicit
constraint tracking, and a final response whose answer items can be checked
against cited webpage sources. The separate taxonomy-development set appears
in Appendix~\ref{app:taxonomy_development}.

\subsection{Operational Task Definition}
\label{app:task_definition}

A retained long-horizon web search task requires an agent to search,
open, read, compare, and verify information from public webpages and then
produce a response that can be checked from those webpages. Tasks may require
multiple candidates, fields, sources, or constraint checks; they are not
restricted to a single answer string or a single supporting page.

For every retained task, the final-answer instruction requires each answer item
or answer group to include the corresponding webpage URL or URLs that verify
the information in that group. When an original task does not specify how many
items should be returned, the adapted task specifies an appropriate minimum
count. This makes task-level coverage measurable while preserving group-level
verification: a task can require at least \(N\) valid groups, while each
proposed group must independently satisfy all applicable hard constraints and
source requirements.

The adapted task text, including any added source or minimum-count requirement,
defines the shared task specification for the evaluated models. The evaluation pipeline does
not require a human-authored gold final answer or pre-labeled trajectory errors
to generate the reported outcome and process diagnostics.

\subsection{Source Composition}
\label{app:task_sources}

The 100-task main evaluation set contains 46 tasks derived from Mind2Web~2,
20 from LiveResearchBench, 29 from AssistantBench, and 5 manually curated
tasks.
Model comparisons aggregate over the complete task set.

The five manually curated tasks are subjected to the same task-definition,
source-grounding, and weak-tool-suitability requirements as the benchmark-derived
tasks. They are included to supplement the retained benchmark subsets rather
than to define a separate evaluation regime.

\subsection{Selection and Adaptation}
\label{app:task_screening}\label{app:task_adaptation}
Selection used five qualitative criteria: public-web feasibility, a need to
compare multiple pages or candidates, explicit jointly applicable constraints,
human-checkable evidence, and compatibility with the five shared tools.
Tasks dependent on login,
maps, decisive visual evidence, or unsupported interaction are excluded.
Adaptation requires supporting URLs per answer group and adds an explicit
minimum count where the original request leaves quantity open. Temporal,
scope, source, formatting, and candidate-level conditions remain in the task
shown to every model; they are not hidden scoring requirements.

\subsection{Stored and Derived Task Metadata}
\label{app:task_metadata}

The core stored task object contains
task\_id, category, and question.
The category field retains heterogeneous descriptive tags. Primary model
results aggregate over all task IDs using the same task population.

\suppEntry{Stored fields}{
task\_\allowbreak id joins the task, trajectory, manifest, Stage~1, and Stage~2
artifacts; question records the adapted task text;
and category is retained only as an unconsolidated descriptive tag.
}
\suppEntry{Derived diagnostics}{
question\_\allowbreak word\_\allowbreak count records prompt length, while
rubric\_\allowbreak variation
describes field-level variation across run-specific rubrics in
Appendix~\ref{app:stage1_verification}.
Neither quantity enters scoring.
}

Because Stage~1 reconstructs a rubric independently for every model--task run,
exact free-text rubric hashes are not treated as task-level ground truth.
All reported outcome scores use the rubric stored in the corresponding run
report.

\subsection{Evaluated Model Inventory}
\label{app:model_inventory}

The retained main comparison comprises 16 models from eight families, each
evaluated on the same 100 tasks with the shared weak-tool controller.
The run-selection manifest binds each retained trajectory to its evaluated
model; \textbf{trace02} identifies DeepSeek-V3.2 in this comparison.

\begin{table*}[tbp]
\centering
\suppTableSetup
\begin{tabularx}{\linewidth}{@{}L{0.14\linewidth}Y r@{}}
\toprule
\textbf{Family} & \textbf{Evaluated models} & \textbf{Trajectories} \\
\midrule
GPT & GPT-6.1-Sol; GPT-6-Luna-Pro; GPT-5.4; GPT-5-Mini & 400 \\
Gemini & Gemini-3.1-Pro; Gemini-2.5-Flash & 200 \\
Kimi & Kimi-K2.6; Kimi-K2.5 & 200 \\
Qwen & Qwen3.7-Plus; Qwen3.6-Flash & 200 \\
DeepSeek & DeepSeek-V3.2; DeepSeek-V4-Flash & 200 \\
GLM & GLM-5; GLM-4.7-Flash & 200 \\
Grok & Grok-4.20 & 100 \\
Nemotron & Nemotron-3-Ultra & 100 \\
\midrule
\textbf{Total} & \textbf{16 models, eight families} & \textbf{1,600} \\
\bottomrule
\end{tabularx}
\caption{Main evaluation inventory across 100 tasks. Each of the 16 models
contributes one retained trajectory per task, giving 1,600 trajectories
across eight families. Model names follow the main paper.}
\label{tab:model_inventory}
\end{table*}

\subsection{Evaluator Role Assignment}
\label{app:evaluator_roles}

Stage~1 uses DeepSeek-V4-Pro as a single verifier for rubric construction and
source-constrained answer verification. Stage~2 uses DeepSeek-V4-Pro as
Judge~A and Qwen3.7-Max as Judge~B; Qwen3.7-Max also supplies the arbiter.
These assignments are fixed across evaluated models and
tasks. Evaluated-model identity is excluded from evaluator inputs.
Stage~1 uses the task, answer groups, and evidence from agent-provided URLs
without replacement search. Each Stage~2 judge receives the task, run-specific
rubric, final response, Stage~1 verdicts, and normalized round, without web
access. Only the arbiter additionally receives the two judge outputs.
Human review validates this automatic framework and is not an operational
step in its evaluation pipeline.

%% file: supplementary_sections/03_collection.tex
\section{Controlled Weak-Tool Trajectory Collection}
\label{app:weak_tool_collection}

This section expands the controlled collection protocol summarized in the main
paper. All evaluated models use the same ReAct controller, system prompt,
five-tool interface, iteration budget, and execution policy; only the
underlying LLM varies during collection. Under these fixed conditions, the
observable trajectory exposes model-side search planning, URL selection, page
reading, keyword localization, link navigation, evidence checking, and final
synthesis. Scoring uses a fixed evaluator configuration with independently
reconstructed run-specific rubrics, as detailed in
Appendix~\ref{app:stage1_verification}.

Retained-run and recovery rules are specified in
Appendix~\ref{app:retained_trajectory_policy}.

\subsection{Measurement Objective and Information Boundary}
\label{app:weak_tool_objective}

The controlled interface keeps page selection, passage inspection, and
strategy revision observable.

\Needspace{7\baselineskip}
\begin{contractbox}{Weak-tool information boundary}
\textbf{Discovery.} web\_search exposes ranked titles, URLs, and domains;
open\_url creates a cached page record and returns metadata, not body text.
\textbf{Evidence.} Only read\_page and find\_on\_page expose cached
page text to the model; extract\_links exposes only anchors and destination
URLs. \textbf{Excluded assistance.} The interface supplies no query-aware
snippets, page summaries, semantic retrieval, reranked passages, direct
answers, or LLM-generated webpage extraction.
\end{contractbox}

\subsection{Shared ReAct Controller}
\label{app:react_controller}

The agent is implemented as a text-style ReAct controller connected to an
OpenAI-compatible model endpoint. Function-calling mode is disabled so that the
controller emits an observable sequence of model messages, tool requests, tool
results, and a terminal final-response marker. These records are captured by the
collection framework and later converted into the normalized trajectory
representation described in Appendix~\ref{app:trajectory_construction}.

A controller iteration may produce a tool request, an observation-dependent
follow-up decision, or a terminal response. Malformed actions are first handled
by the implementation's mechanical parser recovery. If parsing still fails, the
format error and expected action format are returned to the model, and the
correction consumes another ReAct iteration rather than a separate
malformed-action budget.

\subsection{Weak-Tool Interface}
\label{app:weak_tool_api}

The agent registers exactly five web tools. Their roles are intentionally
separated so that discovering a URL, opening a page, reading its body text,
locating a literal term, and following links remain distinct model decisions.
The model-visible contracts are:

\suppEntry{web\_search}{
Input is a query and requested result count, capped at 10. The model receives
only the query plus ranked titles, URLs, and domains. Snippets, page content,
answer boxes, knowledge-graph fields, summaries, semantic chunks, relevance
scores, and direct answers are excluded.
}
\suppEntry{open\_url}{
Input is a candidate URL. The tool returns a page identifier and metadata:
requested and resolved URLs, status, title, cleaned-text length,
extracted-link count, and cache status. It does not return body text, a
relevant passage, a main-content summary, a query-aware fragment, or an
LLM-generated description.
}
\suppEntry{read\_page}{
Inputs are a page identifier, start offset, and requested character count
(default 3,000; hard cap 5,000). The return is an exact character slice from
mechanically cleaned cached text. No rewriting, compression, semantic
filtering, reranking, or summarization is applied.
}
\suppEntry{find\_on\_page}{
Inputs are a page identifier, literal keyword, maximum matches, and context
width. The model receives case-insensitive exact-substring matches with local
cached-text context, without semantic search, BM25, embeddings, query
expansion, fuzzy retrieval, or LLM extraction.
}
\suppEntry{extract\_links}{
Inputs are a page identifier and maximum link count. The tool mechanically
returns anchor text and absolute URLs, without relevance ranking,
recommendations, target-page summaries, or preselection of ``useful'' links.
}

\subsection{Representative ReAct Search Episode}
\label{app:react_search_episode}\label{app:trajectory_collection_case}

\begin{figure*}[!t]
\centering
\includegraphics[
  width=0.76\textwidth
]
{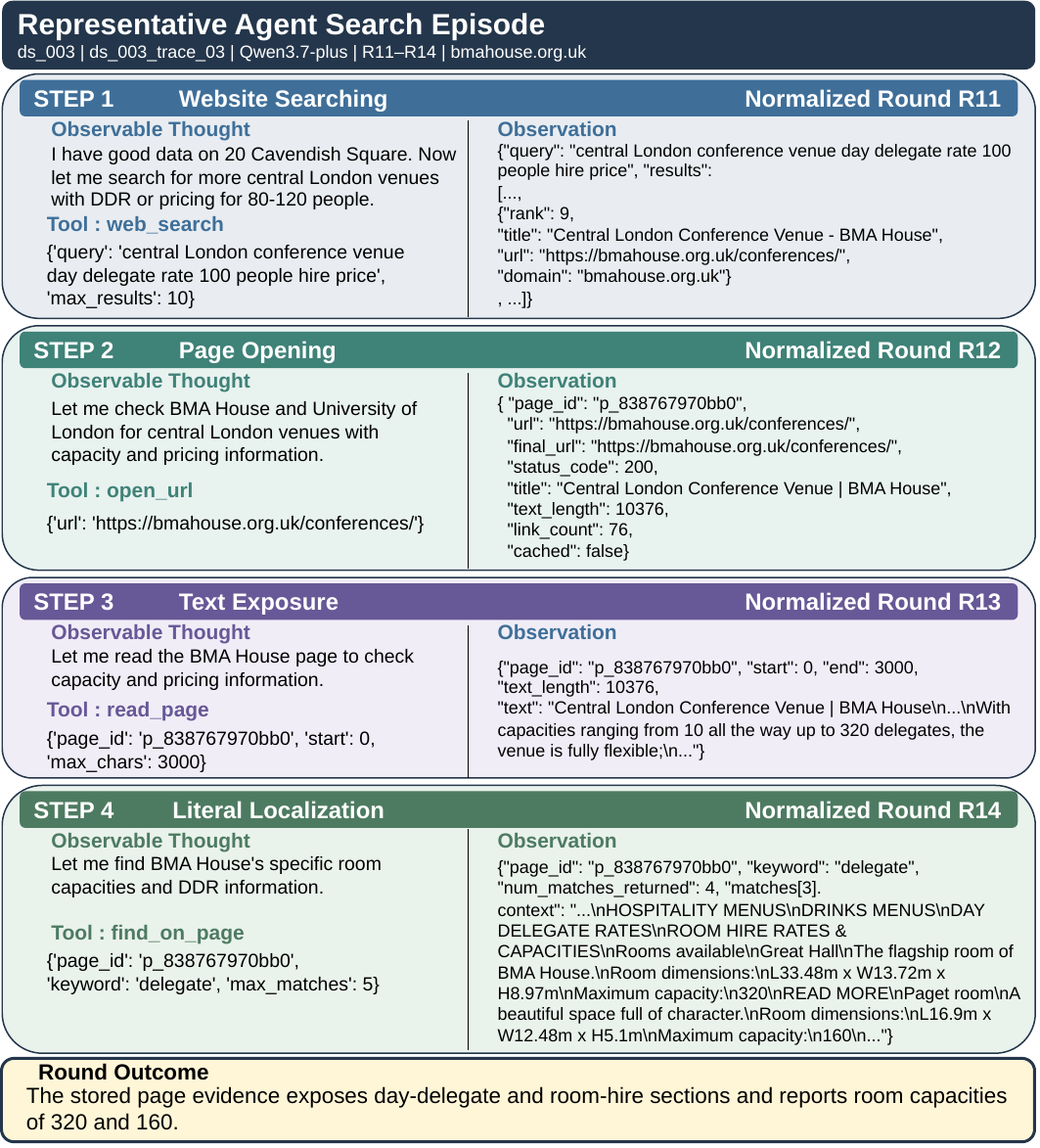}
\Description{Four recorded agent rounds illustrate search, opening, reading, and literal text localization.}
\caption{Recorded rounds R11--R14: candidate discovery, metadata-only opening,
cached-text exposure, and literal localization. R13--R14 reuse the page identifier
from R12; ellipses abbreviate stored fields.}
\label{fig:representative_react_episode}
\label{fig:app_collection_episode}
\end{figure*}

Figure~\ref{fig:representative_react_episode} follows Qwen3.7-Plus on
\textbf{ds\_003\_trace\_03}: R11 discovers the BMA House URL, R12 opens it,
and R13--R14 read and search the same cached page. Each step is a separate
normalized round. R13 exposes 3,000 of the 10,376 cached characters, and R14
returns four literal matches for \textbf{delegate}. The displayed ``Thought''
is model-emitted controller text. The terminal response is stored separately
from these normalized rounds.

\subsection{Effective Execution Controls}
\label{app:execution_controls}

\textbf{Table~\ref{tab:execution_controls}} records the collection configuration.
The 120-iteration budget and \textbf{generate} early-stopping setting are
present in all 1,600 retained run manifests. Other operational values below
come from the archived configuration audit and implementation snapshot;
per-run historical environment overrides are not fully persisted.

\begin{table*}[tbp]
\begin{minipage}{\textwidth}
\centering
\suppTableSetup
\begin{tabularx}{\linewidth}{@{}YL{0.38\linewidth}@{}}
\toprule
\textbf{Control} &
\textbf{Effective value} \\
\midrule

Maximum ReAct budget & 120 iterations \\
Early-stopping method & generate \\
web\_search results & Default 10; hard cap 10 \\
read\_page characters & Default 3,000; hard cap 5,000 \\
LLM request timeout & 600 seconds per attempt \\
web\_search timeout & 10-second connect; 20-second read \\
open\_url timeout & 10-second connect; 25-second read \\
Cached-page tools & No network request \\
End-to-end wall-clock limit & None \\
No-output watchdog / check / grace & 660 / 5 / 5 seconds \\
Search-stagnation thresholds & Soft 8; hard 14 \\
Total-search thresholds & Soft 50; hard 75 \\
Hard-brake evidence target & 3 evidence items \\

\bottomrule
\end{tabularx}
\captionof{table}{
Recorded collection controls; historical per-run coverage is specified in the text. Per-request
timeouts and the subprocess watchdog are distinct from search-stagnation
controls.
}
\label{tab:execution_controls}
\end{minipage}
\end{table*}

Parser corrections consume
ordinary iterations, requests above a tool cap are truncated, and a failed web
request is returned as an observable tool error. The LLM timeout applies per
request attempt, not to the complete trajectory; no fixed end-to-end
wall-clock limit is imposed.

The 660-second watchdog measures absence of subprocess output, not absence of a
new webpage or search result. It is therefore distinct from the soft and hard
search-stagnation counters. The latter monitor search progress inside the agent,
whereas the watchdog monitors whether the batch subprocess is still producing
observable output.

\subsection{Stored Artifacts and Behavioral Diagnostics}
\label{app:collection_artifacts}

For each retained model--task trajectory, the collection project stores raw and
human-inspectable interaction records, run metadata, behavioral metrics, and
checkpoint state. Run identifiers join these records to tasks and model metadata.

\suppEntry{Interaction records}{
The raw record preserves ordered model messages, tool calls, observations, and
terminal output. The human-inspectable trajectory adds normalized
model-decision blocks and trajectory-round boundaries for downstream auditing.
}
\suppEntry{Join and recovery metadata}{
Manifests store run, task, model, trajectory, configuration, and artifact
identifiers; checkpoint state supports resumption without creating an
additional experimental sample.
}
\suppEntry{Behavioral and finalization diagnostics}{
Tool-use fields summarize calls, domains, URLs, and text read. Separate
finalization fields record answer provenance, termination, evidence use,
validator status, premature answering, and budget exhaustion; they do not form
one mutually exclusive state.
}

Existing code identifiers may retain the legacy term trace; the
paper-facing description uses \emph{trajectory}.

The large internal webpage caches are not part of the paper-facing package.
The archived analysis records contain the URLs and observable webpage text presented within the
trajectory records, together with final responses, manifests, metrics, and
evaluation reports. The implications for exact re-aggregation and live-web
re-verification are discussed in Appendix~\ref{app:reproduction}.

%% file: supplementary_sections/04_normalization.tex
\section{Trajectory Construction, Finalization, and Recovery}
\label{app:trajectory_construction}

The collection framework records low-level observable events, whereas
\methodabbr{} operates on normalized trajectory rounds and a separate
terminal final response. This section specifies the transformation between
these representations and distinguishes trajectory construction from
terminal validation, collection-time final synthesis, and infrastructure
recovery.

\subsection{Observable Event Record}
\label{app:observable_events}

The raw record preserves observable controller events in chronological order:
the task question, model-produced ReAct messages, tool requests and results,
and a terminal final-answer marker. The formatted record uses event classes
such as Question, ReasoningRound, ToolCall, ToolCallResult, and FinalAnswer.
Trajectory construction uses only this recorded interaction; hidden model
states and external web evidence are outside the extractor's inputs.

\subsection{Trajectory-Round Construction}
\label{app:trajectory_round_construction}

The production extractor preserves event order and applies deterministic
boundaries to the observable ReAct sequence. A new model-decision turn opens
a round; tool actions and results follow the implemented tool-stage order.
A round may begin at \textbf{open\_url} or a later stage when earlier stages
are absent. A transition back to an earlier stage can open another round:
for example, \textbf{find\_on\_page} followed by \textbf{open\_url} creates
a boundary under the production ordering.

Rounds therefore differ from raw events and individual tool calls. Their
boundaries follow recorded decisions and tool-stage transitions without
post-hoc human segmentation by topic, candidate set, or semantic similarity.
The terminal response is extracted separately and receives no round index.
The production extractor's event-type branches specify the exact boundary
rules.

\subsection{Worked Round Boundaries}
\label{app:round_examples}
The production boundary rules can be read from three retained examples:
\begin{itemize}
\item \textbf{ds\_009\_trace\_03}: new model decisions open R1--R4;
R1--R3 contain search results and R4 an opening result. A read request present
only as unexecuted call text in R4 contributes no page-text observation.
\item \textbf{ds\_058\_trace\_03}: R3 begins directly with an opening result;
the extractor does not invent a preceding search stage.
\item \textbf{ds\_065\_trace\_06}: an opening result in R4 is followed by a
search result, resetting the tool-stage order and starting R5.
\end{itemize}
The final response remains separate in each example.

\subsection{Terminal Final-Response Separation}
\label{app:final_response_separation}

The normalized trajectory is represented as
\[
\begin{aligned}
\tau_{m,q}
  &= \left(r_{m,q,1},\ldots,r_{m,q,T_{m,q}}\right),\\
A_{m,q}
  &= \left(\tau_{m,q},F_{m,q}\right).
\end{aligned}
\]
Stage~1 evaluates \(F_{m,q}\), while Stage~2 audits the indexed rounds
of \(\tau_{m,q}\) conditioned on that response and its Stage~1 judgments.

The terminal-output parser distinguishes a canonical agent final response from
intermediate tool syntax or malformed terminal content. Final-answer validity
checks reject, among other cases:

\begin{itemize}
    \item empty or degenerate repeated output;
    \item raw tool-call text, tool JSON, or XML-style tool-call syntax;
    \item page metadata presented as if it were the answer;
    \item malformed terminal content that cannot be interpreted as a final
    response;
    \item an answer produced before any executed page-text reading or literal-search evidence operation.
\end{itemize}

The last condition implements the evidence-before-answer requirement: search
titles and opened-page metadata alone do not count as page-text evidence.

\subsection{Collection-Time Final-Only Synthesis}
\label{app:collection_finalization_recovery}

If terminal output is invalid but the trajectory already contains usable
evidence, the collection framework may request a no-tool final-only synthesis.
The model receives the already collected context but cannot invoke additional
web tools. The current configuration permits up to two final-synthesis attempts.

A synthesized final answer is recorded through explicit final-answer
provenance fields rather than inferred from answer text or stored as a second
trajectory. The collection record sets final\_answer\_source to
final\_recovery; the manifest and finalization record persist the same
source value. Downstream audit code derives the
final\_only\_synthesis flag strictly from this field.

If usable evidence is unavailable, the framework preserves an invalid fallback
rather than fabricating a source-grounded response. Final-only synthesis
records a distinct terminal-response construction path; the resulting answer
undergoes the same collection checks and subsequent evaluation.

Both evaluation stages receive the retained response unchanged.
Evaluation-time normalization applies to trajectory structure, URLs,
schemas, and judge outputs.

\subsection{Retry and Recovery Layers}
\label{app:retry_layers}

The recovery layers may overlap, but none creates an additional experimental
replicate:

\begin{itemize}
    \item \textbf{Request layer.} Web-tool HTTP calls are not automatically
    retried; a backend exception becomes an observable tool error. Model
    requests use framework retry setting 5, while malformed provider response
    objects permit up to three shape-recovery attempts.
    \item \textbf{Controller layer.} Mechanical parser recovery is attempted
    for malformed ReAct actions; unresolved format errors consume ordinary
    ReAct iterations rather than a separate retry budget.
    \item \textbf{Process layer.} A failed batch subprocess permits up to two
    attempts, and a saved runtime checkpoint permits up to three auto-resume
    attempts. Resume markers remain observable in the retained artifact.
    \item \textbf{Terminal layer.} Collection-time no-tool final-only synthesis
    permits up to two attempts using already collected evidence.
\end{itemize}

Provider connection, timeout, rate-limit, and internal-server exceptions are
handled by the model-request layer. Web-tool exceptions are instead returned as
observable tool errors. If a runtime exception persists, the framework saves a
checkpoint before applying the configured resume policy.

\subsection{Execution-State Dimensions}
\label{app:execution_state_dimensions}

Execution and finalization fields arise from different dimensions and should
not be interpreted as one mutually exclusive status variable.

\suppEntry{Availability and provenance}{
has\_\allowbreak final\_\allowbreak answer records whether a terminal
object is present.
final\_\allowbreak answer\_\allowbreak source distinguishes
agent, final\_\allowbreak recovery, and fallback,
while final\_\allowbreak answer\_\allowbreak reason records the
selected terminal path.
}
\suppEntry{Budget, evidence, and validation}{
max\_\allowbreak iterations\_\allowbreak reached may coexist with a
saved response;
evidence\_\allowbreak requirement\_\allowbreak met records whether
page text was obtained before finalization; and the validity flags plus
final\_\allowbreak answer\_\allowbreak invalid\_\allowbreak reason
record the terminal validator outcome.
}
\suppEntry{Behavior and recovery}{
premature\_\allowbreak answer marks answering before the required page-text
operation. A resume marker records checkpoint-based recovery, although the
released artifacts do not reliably separate every batch retry from every
runtime auto-resume.
}

\subsection{Retained-Trajectory Policy}
\label{app:retained_trajectory_policy}

The corpus contains one retained trajectory for each of the
\(16\times100=1{,}600\) model--task pairs. All 1,600 artifacts are preserved and
evaluated. Infrastructure retries, resume attempts, invalid-final retries, and
finalization repair do not add extra rows to the experimental run matrix.
Final-answer provenance, terminal validation, evidence status, and resume
markers remain separate diagnostic dimensions and may overlap. Their
independently observable corpus counts are reported with the extended results
in Appendix~\ref{app:extended_results}.

Cache use and the procedures for offline aggregation, evaluator re-execution,
and live-web verification are specified in Appendix~\ref{app:reproduction}.

%% file: supplementary_sections/05_stage1.tex
\section{Stage~1: Source-Constrained Outcome Verification}
\label{app:stage1_verification}

Stage~1 evaluates the terminal response as a set of source-grounded answer
groups and assigns one of three verdicts to each extracted group.

\begin{contractbox}{Stage~1 evidence contract}
\textbf{Receives.} The task question and retained terminal response; the
run-specific rubric and answer groups are reconstructed within Stage~1.
\textbf{May access.} Only webpages cited in the evaluated agent's answer-group
URL sets, using saved or newly fetched text for those same URLs.
\textbf{Must not use.} Evaluated-model identity, replacement search, uncited
pages, or external evidence introduced to repair a missing citation.
\textbf{Access-only state.} Unverifiable is reserved for a non-empty cited URL
set whose members all fail solely because of access uncertainty, with
no established answer-visible blocking failure.
\end{contractbox}

\subsection{Inputs and Run-Specific Rubrics}
\label{app:stage1_inputs}

For each retained model--task run $(m,q)$, Stage~1 receives the task question
$q$ and the terminal response $F_{m,q}$. Under the same evaluator configuration,
it independently constructs from the task question
\[
R_{m,q}
=
\left(
N_{m,q},
\mathcal{C}_{m,q},
\rho_{m,q}
\right),
\]
where $N_{m,q}$ is the required number of answer groups under the stored count mode,
$\mathcal{C}_{m,q}$ is the set of hard constraints, and $\rho_{m,q}$ specifies
how the response should be divided into groups. Task-level count requirements
are used for coverage scoring, whereas group-level constraints determine
whether an individual proposed group is acceptable.

The rubric is reconstructed separately for every retained run rather than
generated once and shared across all 16 models. This reflects the implemented
evaluation pipeline; each run is scored using the rubric stored in its own
report. The evaluated model identity is excluded from rubric generation, and no
canonical rubric is substituted post hoc when computing the reported scores.
The fixed evaluator configuration and saved run-specific rubrics jointly
define the scoring protocol. Field-level consistency across the 16
reconstructions per task is summarized at the end of this section.

\subsection{Answer-Group Extraction}
\label{app:answer_group_extraction}

Given $R_{m,q}$, Stage~1 parses the terminal response into
\[
\mathcal{G}(F_{m,q})
=
\left\{
g_{m,q,i}
=
\left(
a_{m,q,i},
\mathcal{U}_{m,q,i}
\right)
\right\}_{i=1}^{P_{m,q}},
\]
where $a_{m,q,i}$ contains the answer content assigned to group $i$ and
$\mathcal{U}_{m,q,i}$ contains the URLs cited for that group. The grouping rule
$\rho_{m,q}$ determines the evaluation unit: depending on the task, a group may
represent one entity, one row, one candidate with several required fields, or
another task-defined answer unit.

Group extraction is automatic and is not manually corrected before scoring.
Multiple URLs may be attached to one group, including cases in which different
URLs support different required fields. Missing citations yield an empty URL
set rather than external source discovery by the evaluator. Responses that do
not follow the requested output format are still parsed when the required
answer units can be recovered from the response; otherwise, missing or malformed
content is reflected in the extracted groups and subsequent verdicts.

\subsection{URL-Constrained Evidence Acquisition}
\label{app:stage1_evidence_acquisition}

For each group $g_i=(a_i,\mathcal{U}_i)$, the verifier may inspect only the URLs
in $\mathcal{U}_i$. It does not issue a search query, replace a missing source,
or obtain additional evidence from an uncited page. Existing usable cached
webpage text is reused. If a cached page is marked abnormal, the evaluator first
attempts to fetch and cache that page again before assigning the source state.

Let $E_i$ denote the webpage evidence obtained from the cited URL set. We set
$\beta_i=1$ when at least one URL yields usable content. We set $\gamma_i=1$
when $\mathcal{U}_i$ is non-empty but every cited URL fails solely because of
access uncertainty, such as a timeout, network failure, anti-bot blocking, or
temporary unavailability. Missing URLs, malformed or fabricated URLs, irrelevant
pages, and reachable pages that do not support the claim are not treated as
access-only failures.

The evaluation-time cache stabilizes verification. Saved reports preserve
the evidence references and verdicts used for statistical re-aggregation
(Appendix~\ref{app:reproduction}).

\subsection{Multiple and Conflicting Sources}
\label{app:stage1_multiple_sources}

When a group cites multiple URLs, the verifier considers their available texts
jointly. Different sources may support different fields or constraints of the
same group, and no single URL is required to establish every field when the
combined cited evidence is sufficient. Conversely, the presence of one relevant
URL does not make the group correct when another required field remains
unsupported.

The verifier does not apply a fixed preference for official, primary, or
secondary sources unless the task itself imposes an explicit source-quality
constraint. When cited sources disagree, the Stage~1 judge resolves the group
using the complete cited evidence and emits one of the existing three verdicts;
the implementation does not introduce a separate unresolved-conflict category.
With mixed usable and inaccessible sources, the verdict follows the usable
cited evidence and applicable answer-visible requirements. The formal rule
below separates these checks from access uncertainty.

\subsection{Group Verdicts}
\label{app:stage1_group_verdicts}

For readability, we suppress the run indices below. Let
$\mathcal{C}_i\subseteq\mathcal{C}$ be the blocking constraints applicable to
group $g_i$, excluding the task-level count requirement. The support function
$\psi(g_i,c,E_i)=1$ indicates support for constraint $c$, checking
cited evidence or answer-visible fields as required by its evidence domain.
Let $\delta_i=1$ when an answer-visible blocking failure is already established,
such as a missing required field or citation. This condition operationalizes
the main paper's access-only principle: an established answer-visible failure
takes priority over page inaccessibility. The group verdict is
\[
v_i =
\begin{cases}
\mathrm{correct}, &
\begin{aligned}
&\beta_i=1\;\wedge\;\delta_i=0
\\[-2pt]
&\;\wedge\;
\psi(g_i,c,E_i)=1
\\[-2pt]
&\;
\forall c\in\mathcal{C}_i,
\end{aligned}
\\[3pt]
\mathrm{unverifiable}, &
\begin{aligned}
&\beta_i=0
\;\wedge\;
\gamma_i=1
\\[-2pt]
&\;\wedge\;\delta_i=0,
\end{aligned}
\\[3pt]
\mathrm{incorrect}, &
\text{otherwise}.
\end{cases}
\]
Thus, missing citations, unsupported claims, invalid attributions, missing
required fields, and constraint violations receive incorrect labels, while
access-only failures are separated as unverifiable.

\subsection{Outcome Metrics and Aggregation}
\label{app:stage1_metrics}

\begingroup
\predisplaypenalty=0
Let $C$, $I$, and $U$ count correct, incorrect, and unverifiable groups,
with $M=C+I+U$. The stored rubric specifies required count $N$ and
count mode $\mu$ (\textbf{count\_mode}). For $M>0$ and $N>0$, define
\[
\begin{gathered}
f(K;N,M,\mu)=
\\[-2pt]
\begin{cases}
\min(K,N)/N, & \mu=\textbf{min},\\
K/\max(N,M), & \mu=\textbf{exact},\\
K/M, & \mu=\textbf{dynamic\_all}.
\end{cases}
\end{gathered}
\]
\endgroup
Then AS $=100f(C;N,M,\mu)$, Upper $=100f(C+U;N,M,\mu)$,
and AP $=100C/\max(M,1)$. Confirmed empty responses receive zero.
Rubrics require $N\geq1$; a zero-denominator branch returns zero.
The minimum-count equation in the main method is the \textbf{min} case;
\textbf{exact} also penalizes overproduction through $\max(N,M)$.
Upper treats unverifiable groups as potentially correct without changing
their labels. Every reported value retains its original run-specific rubric,
quantity mode, and saved automatic score; descriptive modal rubrics do not
replace the stored scoring objects.

AS, AP, and Upper are macro-averages over each model's 100 tasks.
Unverifiable Group Rate instead pools groups:
\[
\mathrm{UGR}_m=100\frac{\sum_q U_{m,q}}{\sum_q M_{m,q}}.
\]
These denominators distinguish response-level precision from pooled
group-level access uncertainty.

\subsection{Rubric Reconstruction Stability}
\label{app:rubric_stability}
\label{app:case_rubric_consistency}

All 1,600 stored rubrics are available, giving 16 reconstructions for each of
100 tasks. Table~\ref{tab:rubric_stability} summarizes the completed
field-consistency analysis. Run-weighted modal coverage sums the largest
within-task frequency once per task and divides by 1,600, including tied modes.
Unique-mode medians exclude only the tasks tied for that field. Each model
score continues to use its own saved rubric; no modal-rubric rescoring was
performed.

\begin{table*}[tbp]
\centering
\suppTableSetup
\begin{tabularx}{\linewidth}{@{}Yrrrr@{}}
\toprule
\textbf{Field} & \textbf{Modal coverage} & \textbf{Tied tasks} &
\textbf{Median unique-mode count} & \textbf{Distinct values/task} \\
\midrule
Required count $N$ & 1,556/1,600 (97.25\%) & 1 & 16/16 (99 tasks) & 1--5 \\
Hard-constraint count & 755/1,600 (47.19\%) & 11 & 7/16 (89 tasks) & 1--9 \\
Normalized grouping-rule text & 927/1,600 (57.94\%) & 2 & 10/16 (98 tasks) & 1--16 \\
\bottomrule
\end{tabularx}
\caption{Field-level agreement among the 16 stored rubrics for each task.
Grouping-rule agreement measures normalized wording; the other rows
measure agreement in the required count and number of constraints.}
\label{tab:rubric_stability}
\label{tab:app_rubric_stability}
\end{table*}

The required count is identical across all 16 runs for 87 tasks, while the
hard-constraint count is identical for one task. Complete normalized rubric
hashes are distinct for every within-task pair (0/12,000 matches).
Normalization applies Unicode NFKC, collapses whitespace, folds case,
sorts dictionary keys and hard constraints, and otherwise preserves list
order. These hashes measure literal agreement in the stored rubric fields.

\begin{figure*}[!t]
\centering
\includegraphics[width=0.93\textwidth]{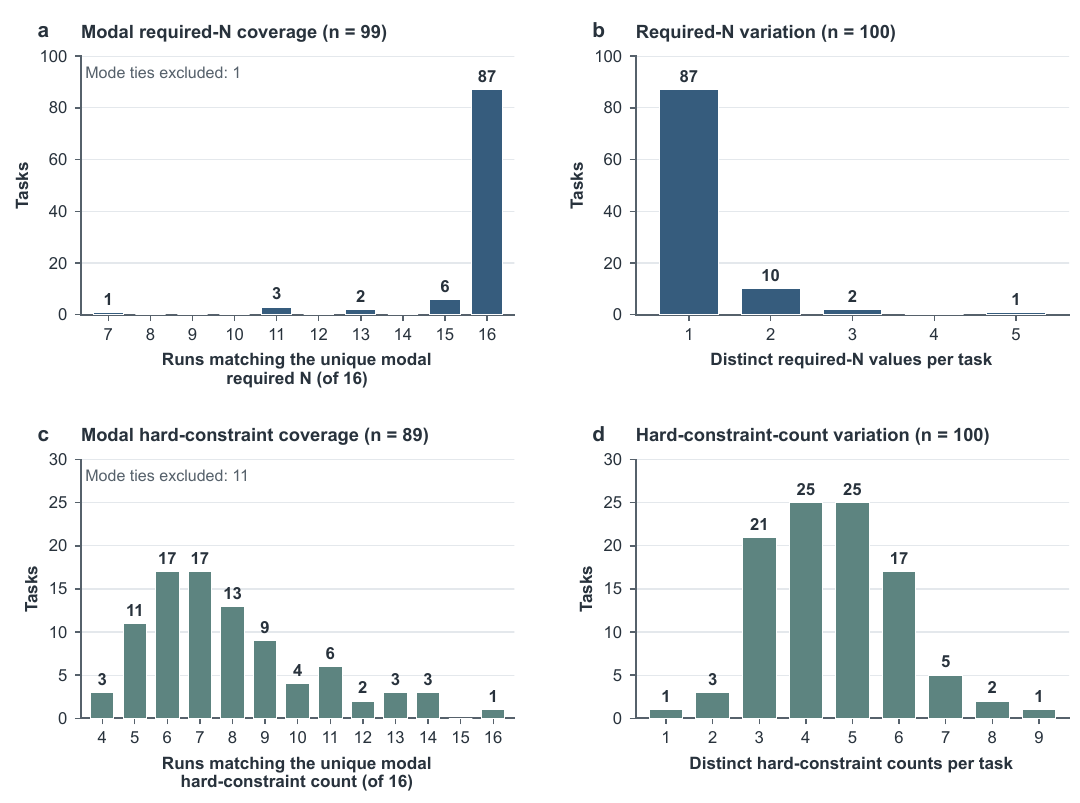}
\Description{Four histograms show required-count and hard-constraint-count stability across sixteen saved rubrics per task. Modal-coverage panels include 99 and 89 tasks after excluding tied modes; distinct-value panels include all 100 tasks.}
\caption{Field-level rubric stability across 16 saved runs per task
(100 tasks; 1,600 rubrics). Panels (a) and (c) show the frequency of the
unique modal required count and hard-constraint count, excluding 1 and 11
tasks with tied modes, respectively. Panels (b) and (d) retain all 100 tasks
and show the number of distinct values per task. The panels measure
structural count agreement. Each run retains its original rubric for scoring.}
\label{fig:supp_rubric_stability}
\end{figure*}

Figure~\ref{fig:supp_rubric_stability} shows the task-level distributions across 16 saved runs per task.

%% file: supplementary_sections/06_stage2.tex
\section{Stage~2: Retrospective Localization and Linking}
\label{app:stage2_localization}

Stage~2 identifies recorded content related to final-answer groups judged
incorrect by Stage~1. It retains the original candidate checks and the stored
\emph{error}/\emph{nonerror}/\emph{skipped} labels. A consensus-\emph{error}
round is called a \emph{failure node}, denoting an automatic candidate flag
under the final-answer context. The output combines this diagnostic inventory
with associations to the original Stage~1 group judgments.

\subsection{Inputs, Verdicts, and Consensus}
\label{app:stage2_audit_inputs}
\label{app:stage2_consensus}
Before judging, a deterministic filter skips structurally empty rounds and
rounds matching configured background or navigation patterns. It applies
lexical rules to the recorded text, context, observations, and URLs, including
price and URL exceptions. Each remaining round is sent to both judges, which
receive the same tuple
\[
x_{m,q,t}=\bigl(q,R_{m,q},F_{m,q},V^{(1)}_{m,q},r_{m,q,t}\bigr),
\]
comprising the task, run-specific rubric, retained final response, Stage~1
group records, and one normalized round. Neither judge has web access,
evaluated-model identity, or hidden model states. The output records the
verdict, failed criteria, proposed group IDs, a reason, and a diagnostic type
when applicable. Extracted answer units and reasons are evaluator
interpretations, stored separately from the model-emitted text.

The candidate checks inspect task-relevant claims, evidence interpretations,
constraint judgments, and candidate content. A \emph{nonerror} verdict means
that no candidate defect is flagged under these checks; \emph{skipped} means
that no evaluable answer unit is retained. These labels support
outcome-conditioned inspection of the recorded content.

\suppEntry{Agreement on error}{
Retain Judge~A's base record; union failed criteria and deduplicate the
proposed-link union. Use A's normalized type when available, otherwise B's.
}
\suppEntry{Agreement on nonerror or skipped}{
Retain Judge~A's record.
}
\suppEntry{Verdict disagreement}{
Pass the original input and both judge records to the arbiter and adopt
its returned record. Type disagreement alone preserves the agreed verdict.
}

\subsection{Round Accounting}
\label{app:stage2_round_flow}
\begin{figure*}[!t]
\centering
\includegraphics[width=0.94\textwidth]{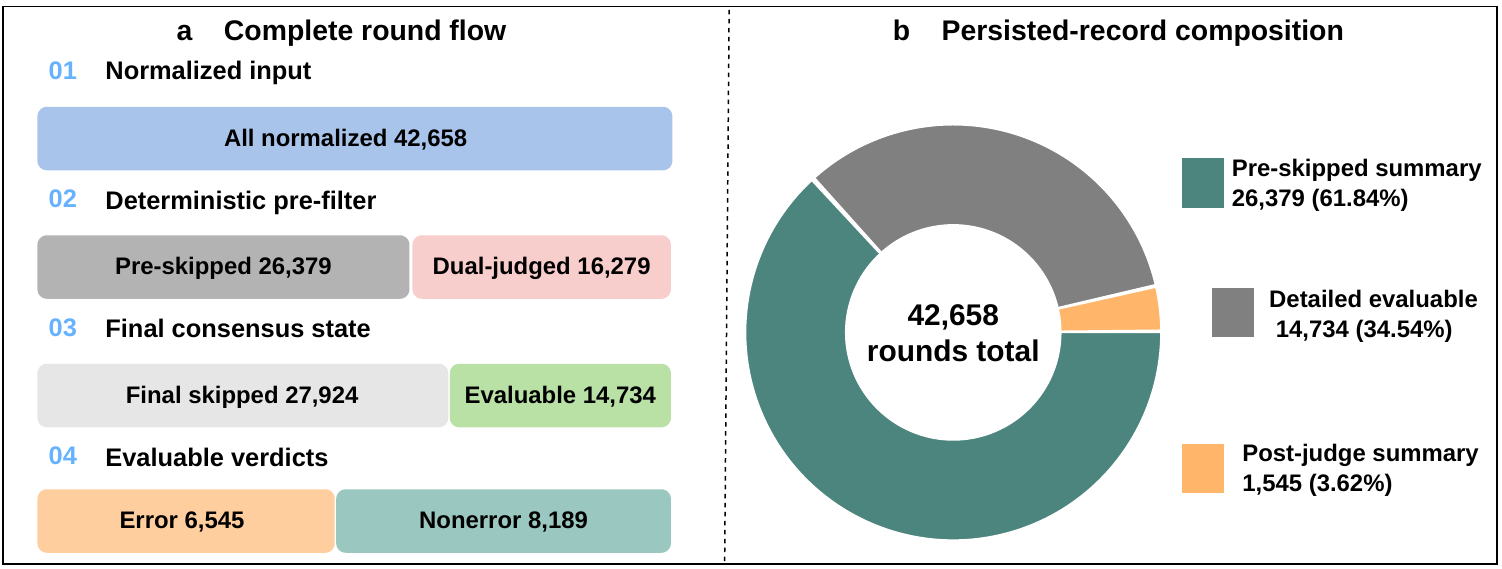}
\Description{Two panels show the complete Stage 2 round flow and persisted-record composition across 42,658 rounds. Pre-skipped, detailed evaluable, and post-judge skipped records number 26,379, 14,734, and 1,545, respectively.}
\caption{Stage~2 round accounting across 1,600 runs. (a) The 16,279
dual-judged rounds yield 14,734 evaluable records and 1,545 post-judging skips;
the latter combine with 26,379 pre-skipped rounds to give 27,924 final skips.
The stored error label denotes an automatic candidate flag.
(b) Persisted-record composition uses all 42,658 rounds as its denominator:
61.84\% pre-skipped summaries, 34.54\% detailed evaluable records, and
3.62\% post-judge summaries.}
\label{fig:supp_stage2_round_flow}
\end{figure*}

Figure~\ref{fig:supp_stage2_round_flow} accounts for all 42,658 normalized
rounds. Deterministic pre-filtering skips 26,379 rounds before judging;
16,279 proceed to both judges. Of these, 1,545 are skipped after judging,
leaving 14,734 evaluable records: 6,545 candidate flags and 8,189 nonerror
verdicts. Skipped records are compacted into summaries. These counts concern
rounds, rather than provider requests or batches.

\subsection{Judge Agreement and Record Coverage}
\label{app:internal_consistency}

The paired-record diagnostics reported in the main paper cover 14,818
recovered Judge~A/B verdict pairs: 12,417 agree (83.80\%) and 2,401 differ.
Of 4,679 pairs jointly labeled error, 4,612 have comparable normalized types;
4,032 type labels agree (87.42\%) and 580 differ. The remaining 67 pairs
contribute to verdict-level accounting and retain a separate
non-comparable-type status.

Arbiter-return summaries cover 2,317 disagreements; another 84 disagreements
have no retained summary. These two counts sum to the 2,401 verdict
disagreements. Paired-record rates use rounds for which both judge objects
were recovered, including recoverable post-judging skipped records. The
16,279 dual-judged rounds in Appendix~\ref{app:stage2_round_flow} instead
describe the complete operational workload. Each statistic therefore retains
its corresponding record population.

\subsection{Proposed Links and Target-Status Filtering}
\label{app:stage2_linking}
The prompt supplies all Stage~1 groups and their verdicts and asks the judges
to consider task-relevant correspondences such as entities, names, URLs,
providers, prices, dates, and required fields. Exact lexical overlap is not
required. Parsing normalizes ID shape, including null-to-empty-list conversion and
numeric-to-string conversion. Returned-link feature correspondence is assessed in the
human-validation sample.
Agreeing-error links are unioned and deduplicated. Arbiter-resolved links
remain as returned before summary construction.

Let $\bar{\mathcal L}_t$ be the stored consensus links and
$\mathcal B=\{i:v_i=\mathrm{incorrect}\}$ the Stage~1-incorrect group set.
Downstream summaries retain
\[
\mathcal L_t^{\mathrm{sum}}=\bar{\mathcal L}_t\cap\mathcal B.
\]
This intersection excludes correct, unverifiable, and nonexistent targets.
This operation selects targets according to their saved Stage~1 status.
Returned-link feature correspondence is evaluated through the human link audit
in Appendix~\ref{app:human_validation}.

\subsection{Diagnostic Units and Interpretation}
\label{app:stage2_diagnostics}
For a retained run, define
\[
\begin{aligned}
\mathcal E_{\mathrm{all}}&=\{t:\bar z_t=\mathrm{error}\},\\
\mathcal E_{\mathrm{linked}}&=\{t\in\mathcal E_{\mathrm{all}}:
                  \mathcal L_t^{\mathrm{sum}}\ne\emptyset\},\\
t_{\mathrm{first}}&=\min\mathcal E_{\mathrm{linked}}
                  \quad(\mathcal E_{\mathrm{linked}}\ne\emptyset),\\
\mathcal B_{\mathrm{unlinked}}&=\mathcal B\setminus
                  \bigcup_{t\in\mathcal E_{\mathrm{all}}}\mathcal L_t^{\mathrm{sum}}.
\end{aligned}
\]
The primary retrospective output is the linked nodes and their retained
round--group edges; the full candidate inventory also includes unlinked
flags. The earliest linked index identifies the earliest \emph{returned
related location} in the retained trajectory. Nodes, edges, target groups,
and runs have separate denominators: one node may link to multiple groups,
and one group may receive multiple links. Node counts summarize automatic
flags; human link review measures feature correspondence, and the separate
commitment-based audit assesses explicit erroneous commitments.

\subsection{Unassigned Types}
\label{app:stage2_unassigned_types}
Of 6,545 flagged nodes, 6,503 have a normalized type and 42 (0.64\%) remain
unassigned. These nodes stay in localization and linking totals but are
kept separate from the typed-node denominator. The ten-category composition
appears in Appendix~\ref{app:error_taxonomy}.

%% file: supplementary_sections/07_taxonomy.tex
\section{Error-Type Definitions and Representative Cases}
\label{app:error_taxonomy}

\begin{table*}[tbp]
\centering
\suppTableSetup
\begin{tabularx}{\linewidth}{@{}L{0.22\linewidth}Y@{}}
\toprule
\textbf{Error type} & \textbf{Diagnostic definition and nearest boundary} \\
\midrule
\textbf{Wrong entity} & The target is identified, substituted, or merged with a
different entity, alias, version, institution, product, or location. If the
entity is the intended one but fails an eligibility condition, use constraint violation. \\
\addlinespace[3pt]
\textbf{Constraint violation} & A candidate fails an explicit hard condition such as
a feature, threshold, status, or eligibility rule. Date, geographic, and
population boundaries belong to temporal or scope error. \\
\addlinespace[3pt]
\textbf{Missing required field} & An identifiable answer unit lacks a required
attribute or source. An existing value assigned to the wrong field is a
mapping error; incorrectly formed units are a grouping error. \\
\addlinespace[3pt]
\textbf{Insufficient evidence} & The available observation is too incomplete to
establish a claim, for example a title, metadata, or partial passage lacking
the required detail. \\
\addlinespace[3pt]
\textbf{Source not supporting} & Inspected content can be assessed but is irrelevant,
omits the claimed fact, supports another value, or contradicts the
interpretation. The issue is support for the claim, rather than source type. \\
\addlinespace[3pt]
\textbf{Source quality error} & The source is invalid, inaccessible, unsuitable, or explicitly
disallowed by an official/primary-source requirement. \\
\addlinespace[3pt]
\textbf{Count or grouping error} & The number, distinctness, or boundaries of answer
units are wrong through miscounting, duplication, merging, splitting, or
omission. This differs from an absent attribute in a properly formed unit. \\
\addlinespace[3pt]
\textbf{Temporal or scope error} & Inclusion or exclusion crosses an explicit date,
period, geography, population, domain, jurisdiction, or other scope boundary. \\
\addlinespace[3pt]
\textbf{Formatting or mapping error} & The required information exists but violates the
required format or is attached to the wrong item, field, label, unit, column,
or output slot. \\
\addlinespace[3pt]
\textbf{Other error} & A concrete diagnostic defect outside the nine named
categories. Use only when no named category applies; never for an absent type. \\
\bottomrule
\end{tabularx}
\caption{Definitions and category boundaries of all ten diagnostic categories. Every category is defined here; the four selected cases illustrate
a subset of the vocabulary. Stored labels remain
automatic diagnostic assignments, including when the associated webpage
observation itself is correct.}
\label{tab:taxonomy_compact}
\end{table*}

The diagnostic vocabulary contains ten error-type labels for
flagged candidate content. This section first defines every category and its
boundary, then presents four source--answer comparisons, and finally reports
how the vocabulary is fixed and distributed across the corpus.

\subsection{Ten Categories and Their Boundaries}
\label{app:taxonomy_categories}
\label{app:taxonomy_boundaries}
Table~\ref{tab:taxonomy_compact} gives the complete definitions. Each typed
node receives one dominant category. A retained round--group link, when
present, identifies the associated final-answer content. The four cases below use these same category names.

\input{supplementary_sections/10_cases}

\begin{figure*}[!t]
\centering
\includegraphics[width=0.94\textwidth]{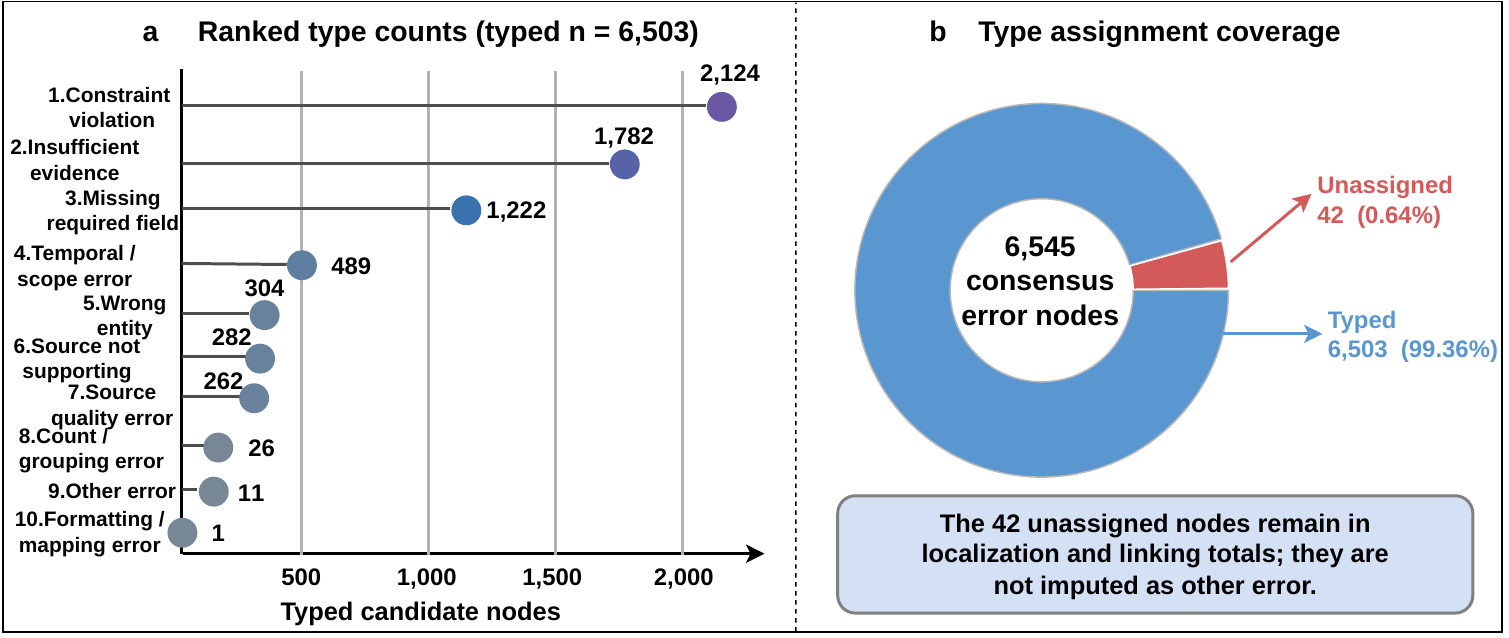}
\Description{Two panels show descending counts for ten diagnostic categories among 6,503 typed candidate nodes, and type assignment coverage among 6,545 candidate flags: 6,503 typed and 42 unassigned.}
\caption{Diagnostic-type composition and assignment coverage.
(a) Counts over 6,503 typed candidate nodes; all ten categories are retained.
(b) Among 6,545 automatically flagged nodes, 6,503 (99.36\%) have a normalized
type and 42 (0.64\%) remain unassigned. The latter stay in localization and
linking totals under a separate unassigned status. Consensus error nodes
denote the automatic candidate flags defined in Appendix~\ref{app:stage2_localization}.}
\label{fig:supp_taxonomy_profile}
\end{figure*}

\subsection{Development and Corpus Profile}
\label{app:taxonomy_development}
\label{app:taxonomy_development_set}
A disjoint development set of 50 tasks and 500 trajectories fixes the
vocabulary used for the 100-task, 1,600-run evaluation.

\label{app:taxonomy_assignment}
For agreeing error verdicts, the consensus retains Judge~A's normalized type
when available, otherwise Judge~B's. Differing verdicts use the arbiter's
returned record. Type disagreement alone is not arbitrated. An absent
normalized type remains unassigned; \emph{other error} is an explicit tenth
category and is not a missing-value code.

\label{app:taxonomy_profile}
Interpretation follows three levels: the recorded webpage observation, the
agent's emitted text, and the evaluator's diagnostic assignment. Inspecting
the linked final-answer group then places the round in the context of the
retained outcome.

The full evaluation retains 6,503 typed and 42 unassigned nodes among 6,545
automatic flags. Constraint violation, insufficient evidence, and missing
required field together account for 5,128/6,503 typed nodes (78.86\%). All ten
categories remain present, including one formatting/mapping node.
Model-specific shares use that model's own typed-node count and exclude its
unassigned nodes. The resulting shares describe the composition of the
automatic diagnostic inventory.

%% file: supplementary_sections/10_cases.tex
\subsection{Four Representative Source--Answer Cases}
\label{app:answer_process_case}
\label{app:qualitative_cases}
\label{app:case_overview}
\label{app:four_case_records}

\begin{figure*}[tbp]
\centering
\includegraphics[width=\textwidth]{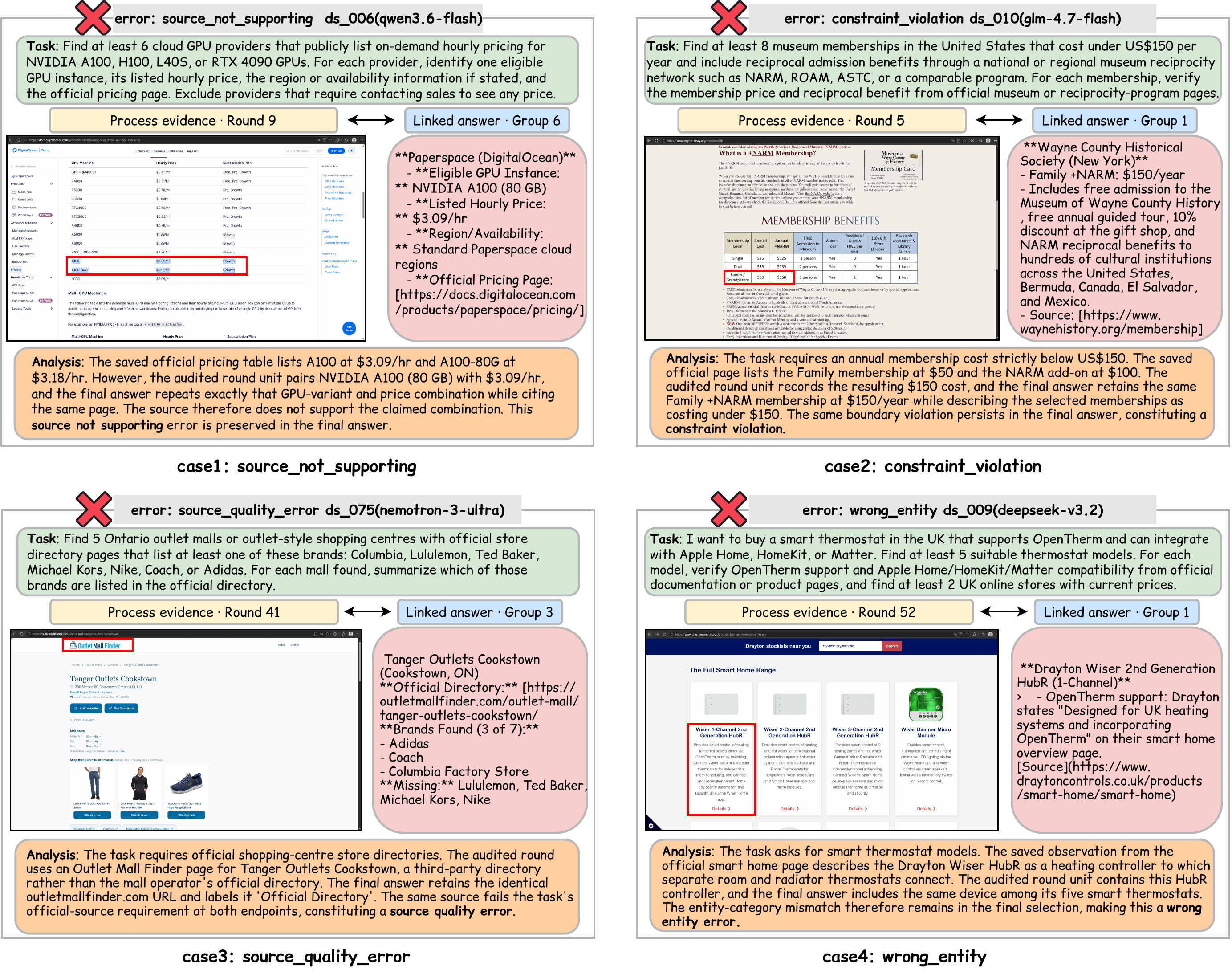}
\Description{Four original panels pair recorded observations with final-answer
 groups: GPU variant pricing, a museum membership price threshold, shopping
 directory provenance, and a heating-controller entity category.}
\caption{Illustrative source--answer comparisons: source not supporting,
constraint violation, source quality error, and wrong entity. The original
four-panel layout pairs returned observations with final-answer groups;
round units and type descriptions reproduce evaluator interpretations.
The accompanying case descriptions identify each model, task, round, and group.}
\label{fig:app_four_cases}
\end{figure*}

Figure~\ref{fig:app_four_cases} illustrates four categories defined in
Table~\ref{tab:taxonomy_compact}: source not supporting, constraint
violation, source quality error, and wrong entity. Each panel pairs a recorded
observation with its final-answer group and the evaluator's interpretation.
The remaining six categories retain their full definitions in the same table.
The cases preserve the original model, task, round, and group bindings:
\begin{enumerate}
\item \textbf{Source not supporting.} Qwen3.6-Flash, \textbf{ds\_006},
R9/G6. The source lists A100 at USD 3.09/hour and A100-80G at USD 3.18/hour;
the answer pairs the 80-GB variant with USD 3.09/hour.
\item \textbf{Constraint violation.} GLM-4.7-Flash, \textbf{ds\_010},
R5/G1. The Family membership and NARM add-on total USD 150/year under a
task requirement of strictly less than USD 150.
\item \textbf{Source quality error.} Nemotron-3-Ultra, \textbf{ds\_075},
R41/G3. A third-party Outlet Mall Finder page is retained as an
``Official Directory'' for Tanger Outlets Cookstown.
\item \textbf{Wrong entity.} DeepSeek-V3.2, \textbf{ds\_009},
R52/G1. The source describes Drayton Wiser HubR as a heating controller;
the final selection includes it among smart thermostats.
\end{enumerate}

These panels provide concrete source--answer comparisons for retained answer
groups; figure-internal round units and type descriptions are evaluator
interpretations. The separate corpus-level human link audit uses the fixed
317-edge sample in Appendix~\ref{app:evaluator_reliability}.

%% file: supplementary_sections/08_results.tex
\section{Extended Results and Diagnostics}
\label{app:extended_results}
\label{app:extended_diagnostics}

\begin{figure*}[tbp]
\centering
\includegraphics[width=\textwidth]{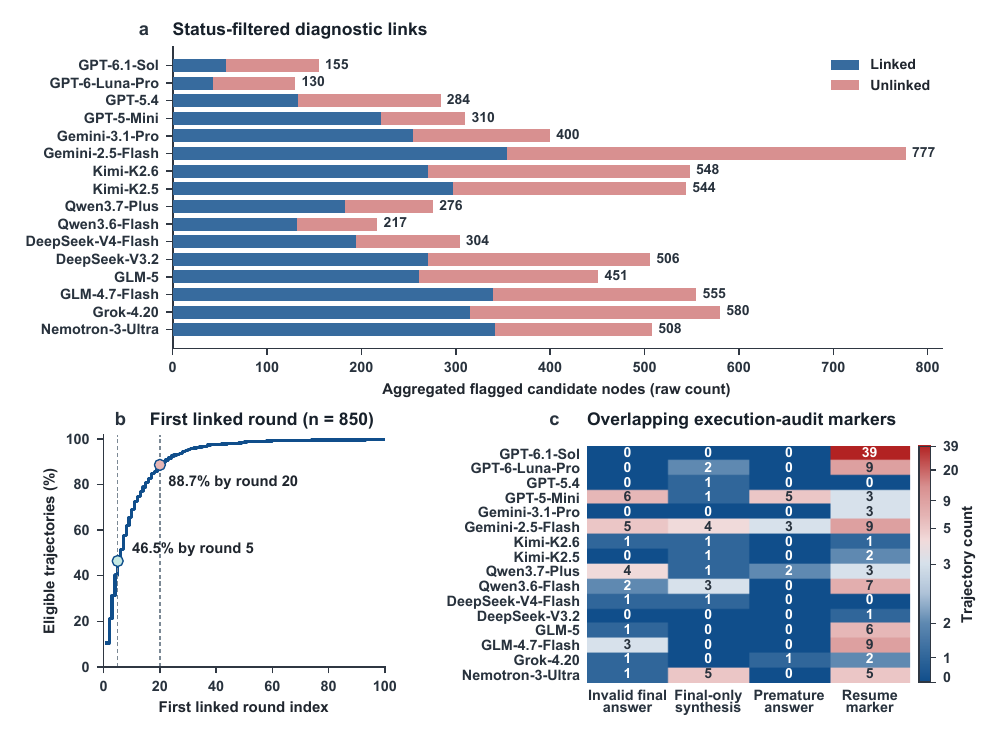}
\Description{Three panels show linked and unlinked candidate-node counts
for sixteen models, the cumulative distribution of earliest returned
associations among 850 linked trajectories, and four overlapping execution
markers for each model.}
\caption{Extended trajectory and execution diagnostics for 1,600 retained
runs. (a) Linked and unlinked candidate flags, including unassigned types.
(b) Earliest returned association among 850 trajectories with links.
(c) Separately recorded, overlapping execution markers. The marker totals
are 25 invalid-final flags, 20 final-only syntheses, 11 premature-answer flags,
and 99 resume-marked trajectories.}
\label{fig:supp_extended_diagnostics}
\end{figure*}

The main comparison covers 16 LLMs from eight families on 100 tasks:
1,600 retained trajectories, 7,685 answer groups, and 42,658 normalized
rounds. Automatic evaluation flags 6,545 failure nodes, comprising 6,503
typed and 42 unassigned nodes. The retained associations comprise 3,664
answer-linked nodes and 5,798 round--group edges. These corpus totals
anchor the model-level and conditional distributions below.
Figure~\ref{fig:supp_extended_diagnostics} gives a joint view of the linked
candidate inventory, earliest returned associations, and execution markers.

\subsection{Outcome Scores and Aggregation}
\label{app:extended_outcome}

The main paper provides the complete model-level outcome table.
AS, AP, and Upper are unweighted means over each model's 100
saved reports. UGR pools unverifiable groups over all provided groups for
that model. AP and UGR are displayed as percentages. The corpus contains
3,863 correct, 3,374 incorrect, and 448 unverifiable groups.
The stored count modes comprise 1,294 \textbf{exact}, 164 \textbf{min}, and
142 \textbf{dynamic\_all} runs. The formulas in
Appendix~\ref{app:stage1_metrics} reproduce all 1,600 saved scores from
their original group counts; no labels or rubrics are changed.

\subsection{Complete Model-Level Process Diagnostics}
\label{app:extended_process}

A linked node is a non-skipped consensus-error round whose proposed group
IDs intersect the Stage~1 incorrect groups. One node may link to multiple
groups; each retained edge is identified by its run, round, and group IDs.
Table~\ref{tab:extended_process_counts} separates these units. Of the
3,374 incorrect groups, 2,229 (66.06\%) have at least one retained association
and 1,145 have none. The 3,664 linked and 2,881 unlinked
nodes partition all 6,545 failure nodes.
These counts include all 42 unassigned nodes, with each node counted once
in Figure~\ref{fig:supp_extended_diagnostics}(a).

\begin{table*}[tbp]
\centering
\suppDenseTableSetup
\begin{tabular*}{\linewidth}{@{\extracolsep{\fill}}lrrrrrr@{}}
\toprule
\textbf{Model} & \textbf{\shortstack{All\\rounds}} & \textbf{\shortstack{Evaluable\\rounds}} & \textbf{\shortstack{Flagged\\nodes}} & \textbf{\shortstack{Runs with\\flag}} & \textbf{\shortstack{Linked\\nodes}} & \textbf{\shortstack{Runs with\\link}} \\
\midrule
GPT-6.1-Sol & 1,774 & 762 & 155 & 43 & 56 & 20 \\
GPT-6-Luna-Pro & 1,841 & 552 & 130 & 49 & 43 & 21 \\
DeepSeek-V3.2 & 2,767 & 1,221 & 506 & 86 & 270 & 54 \\
GPT-5.4 & 1,967 & 838 & 284 & 70 & 133 & 53 \\
DeepSeek-V4-Flash & 3,022 & 853 & 304 & 60 & 194 & 47 \\
Kimi-K2.6 & 3,467 & 1,348 & 548 & 72 & 270 & 50 \\
Kimi-K2.5 & 3,248 & 1,231 & 544 & 76 & 297 & 63 \\
Qwen3.7-Plus & 2,044 & 603 & 276 & 73 & 183 & 56 \\
Grok-4.20 & 2,540 & 1,102 & 580 & 76 & 315 & 55 \\
GLM-5 & 3,004 & 1,081 & 451 & 72 & 261 & 60 \\
GPT-5-Mini & 2,507 & 693 & 310 & 61 & 221 & 54 \\
Nemotron-3-Ultra & 3,569 & 1,122 & 508 & 80 & 341 & 64 \\
Gemini-3.1-Pro & 2,457 & 796 & 400 & 73 & 255 & 62 \\
Qwen3.6-Flash & 2,034 & 425 & 217 & 67 & 132 & 53 \\
Gemini-2.5-Flash & 3,237 & 1,215 & 777 & 79 & 354 & 67 \\
GLM-4.7-Flash & 3,180 & 892 & 555 & 79 & 339 & 71 \\
\textbf{Total} & \textbf{42,658} & \textbf{14,734} & \textbf{6,545} & \textbf{1,116} & \textbf{3,664} & \textbf{850} \\
\bottomrule
\end{tabular*}
\caption{Raw process counts over 100 trajectories per model. Evaluable rounds have an error or nonerror consensus verdict; final skipped rounds are excluded. Linked nodes and runs use the retained intersection with Stage~1 incorrect groups. Counts summarize observed activity and diagnostic coverage within each model's retained trajectories.}
\label{tab:extended_process_counts}
\end{table*}

Failure-node count has pooled Spearman correlations of $-0.427$ with AS
and $-0.428$ with AP ($n=1,600$ trajectories); within-model coefficients
are negative for all 16 models. These are outcome-aware associations:
Stage~2 receives the final response and Stage~1 verdicts.

\subsection{Earliest Returned Association}
\label{app:first_linked_distribution}

Of the 1,600 trajectories, 850 have a retained association. Their first
linked round has median 6 (quartiles 3 and 13; range 1--100). The cumulative
counts are 395/850 by round 5 (46.47\%), 587/850 by round 10 (69.06\%),
and 754/850 by round 20 (88.71\%). The 750 trajectories without an
association remain outside this conditional distribution. These positions
locate the earliest returned related content. Quantiles use linear
interpolation at index $(n-1)q$ in the ordered eligible observations.
Figure~\ref{fig:supp_extended_diagnostics}(b) visualizes this conditional
distribution and reports the cumulative percentages rounded to one decimal.

\subsection{Search Behavior and Outcomes}
\label{app:behavior_outcome}

Table~\ref{tab:search_behavior_means} reports per-trajectory tool-use means. Tool counts measure returned tool-result
events, including unsuccessful returns; they do not count provider retries.
Mean normalized trajectory length ranges from 17.74 to 35.69 rounds.

\begin{table*}[tbp]
\centering
\suppDenseTableSetup
\begin{tabular*}{\linewidth}{@{\extracolsep{\fill}}lrrrrrrr@{}}
\toprule
\textbf{Model} & \textbf{Search} & \textbf{Open} & \textbf{Read} & \textbf{Find} & \textbf{Extract} & \textbf{Tools} & \textbf{Rounds} \\
\midrule
GPT-6.1-Sol & 9.49 & 13.41 & 7.75 & 4.30 & 0.85 & 35.80 & 17.74 \\
GPT-6-Luna-Pro & 14.57 & 9.29 & 5.38 & 4.15 & 0.39 & 33.78 & 18.41 \\
DeepSeek-V3.2 & 13.75 & 15.64 & 10.84 & 12.98 & 0.65 & 53.86 & 27.67 \\
GPT-5.4 & 11.19 & 10.40 & 4.86 & 8.68 & 0.76 & 35.89 & 19.67 \\
DeepSeek-V4-Flash & 21.99 & 14.21 & 12.09 & 5.27 & 0.35 & 53.91 & 30.22 \\
Kimi-K2.6 & 21.98 & 17.42 & 15.64 & 8.59 & 0.32 & 63.95 & 34.67 \\
Kimi-K2.5 & 18.88 & 15.97 & 16.10 & 6.56 & 0.24 & 57.75 & 32.48 \\
Qwen3.7-Plus & 13.80 & 9.63 & 6.87 & 3.82 & 0.16 & 34.28 & 20.44 \\
Grok-4.20 & 10.67 & 11.94 & 14.90 & 4.97 & 0.14 & 42.62 & 25.40 \\
GLM-5 & 13.58 & 19.07 & 13.54 & 6.44 & 0.53 & 53.16 & 30.04 \\
GPT-5-Mini & 16.35 & 10.29 & 2.38 & 9.40 & 0.30 & 38.72 & 25.07 \\
Nemotron-3-Ultra & 13.76 & 23.75 & 11.86 & 9.46 & 0.62 & 59.45 & 35.69 \\
Gemini-3.1-Pro & 16.64 & 10.48 & 8.98 & 5.87 & 0.35 & 42.32 & 24.57 \\
Qwen3.6-Flash & 15.90 & 8.31 & 6.08 & 2.53 & 0.15 & 32.97 & 20.34 \\
Gemini-2.5-Flash & 11.42 & 12.87 & 11.88 & 15.12 & 0.94 & 52.23 & 32.37 \\
GLM-4.7-Flash & 20.85 & 10.83 & 3.87 & 13.79 & 0.12 & 49.46 & 31.80 \\
\bottomrule
\end{tabular*}
\caption{Mean tool use and normalized length over 100 trajectories per model. The Tools column sums the five tool categories; normalized rounds are counted separately from tool calls.}
\label{tab:search_behavior_means}
\end{table*}

Page-use summaries complement these tool counts. Across models, the mean
number of distinct opened URLs ranges from 7.73 to 18.38 per trajectory,
and distinct domains from 4.73 to 9.16. Mean read volume ranges from 4.78 to
48.02 thousand characters, including repeated or overlapping reads; final
answers contain 7.43--14.34 URL occurrences on average. The archived
model-level tables retain the full breakdowns.

Across the eleven recorded behavior fields and the AS/AP outcomes, the
largest absolute pooled Spearman coefficient is 0.140 (final-answer URL
occurrences versus AS). The corresponding AP association is 0.128.
Each pooled coefficient uses 1,600 trajectories; within-model summaries use
100 trajectories per model and report the median of 16 coefficients, with
average ranks for ties. These summaries describe the relation between
observable search activity and the retained outcome scores.

\subsection{Execution-Audit Summary}
\label{app:extended_recovery}
Figure~\ref{fig:supp_extended_diagnostics}(c) shows the model-level
distribution of four execution markers. Each marker counts the trajectories
where it is present; a trajectory can contribute to multiple columns.

The final-answer source field partitions the corpus into 1,567 agent
responses, 20 collection-time final-only syntheses, and 13 fallbacks.
All retained trajectories contain a final answer. The four overlapping
execution markers in Figure~\ref{fig:supp_extended_diagnostics}(c) total
25 invalid-final flags, 20 final-only syntheses, 11 premature-answer flags,
and 99 resume-marked trajectories. The resume indicator counts the literal
marker across batch resumption and runtime continuation.

Collection-time validation records 1,573 valid-final flags and 25
invalid-final flags; two runs have neither flag. An invalid-reason field
is present in 27 records, and 1,589 runs meet the evidence requirement.
Among the 20 synthesis responses, 17 are marked valid and three invalid;
no run has a maximum-iteration flag. These separately stored finalization
fields describe execution checks, while Stage~1 evaluates source-constrained
answer-group correctness. The evaluator preserves the collected answers.

%% file: supplementary_sections/09_validation.tex
\section{Human Validation}
\label{app:evaluator_reliability}\label{app:human_validation}
Human review assesses supplied answer-group labels, response scores, and
returned round--group correspondences. Each analysis uses its specified
sample and denominator.

\subsection{Sample and Review Protocol}
The post-hoc audit uses two randomly selected responses from different
models for each of the 100 tasks: 200 responses and 946 supplied answer
groups, including four explicitly confirmed empty-group responses.
Fifty randomly selected tasks contribute both trajectories to process
review (100 trajectories; 2,642 normalized rounds). Reviewer H1 completes
the primary audit; a different reviewer, H2, independently checks fixed
subsets of 40 responses from 20 tasks (172 groups) and 20 trajectories
from 10 tasks (499 rounds). The subsets form independent agreement checks within the primary samples. Sampling uses CPython's MT19937 generator with seed
20261004 and the frozen 16-model population; it is not stratified by
scores, disagreements, or error counts.

For Stage~1, reviewers receive the requirements, complete final response,
supplied groups, and original cited URLs. Automatic labels, scores,
reasons, separate model metadata, and the other reviewer's annotations
are withheld. Complete answer text may still contain self-identification.
Reviewers assign C/I/U labels and scores are derived from the locked
details; disagreements remain unchanged. Both reviewers self-report
web-search expertise, rule familiarity, and no exposure to the withheld
item-level results. This post-hoc protocol was clarified in the context of
previously inspected comparison results; item-level annotations follow the
withholding and locking procedure described here.

With group boundaries fixed, the audit evaluates supplied-group judgments
and their derived scores. Reviewers check the original citation set at review
time; webpage content may vary between accesses. Each reviewer's annotations are locked before
score comparison. Process labels and the answer endpoints used for link
review are locked separately before candidates are shown, and later
link judgments do not overwrite those endpoints. Process review follows
an instruction to use the current and earlier rounds; the static workbooks
retain the complete trajectories.

Reported 95\% intervals use 10,000 paired task-cluster percentile bootstrap
draws (seed 20261009), preserving all observations and repeated-task weights.
The cluster counts are 100 for primary answers, 50 for process and links,
20 for H2 answers, and 10 for H2 process review. Link resampling includes
the 13 process-sample tasks with no candidate edge.

\subsection{Answer-Group and Score Agreement}
Automatic and H1 labels agree on 894/946 groups (94.50\%, 95\% CI
$[91.96,96.73]$; Cohen's $\kappa=0.8989$, $[0.8523,0.9392]$).

H1 and H2 agree on 160/172 subset groups (93.02\%, $[87.65,97.66]$;
$\kappa=0.8574$, $[0.7464,0.9499]$).

\begin{table*}[tbp]
\centering\suppTableSetup
\begin{tabularx}{\linewidth}{@{}Yrrrr@{\hspace{1.5em}}Yrrrr@{}}
\toprule
\multicolumn{5}{c}{Automatic $\rightarrow$ H1 (946 groups)} &
\multicolumn{5}{c}{H1 $\rightarrow$ H2 (172 groups)} \\
\cmidrule(lr){1-5}\cmidrule(lr){6-10}
Row label & C & I & U & Total & Row label & C & I & U & Total \\
\midrule
C & 475 & 21 & 2 & 498 & C & 103 & 5 & 1 & 109 \\
I & 22 & 373 & 1 & 396 & I & 6 & 51 & 0 & 57 \\
U & 1 & 5 & 46 & 52 & U & 0 & 0 & 6 & 6 \\
\midrule
Total & 498 & 399 & 49 & 946 & Total & 109 & 56 & 7 & 172 \\
\bottomrule
\end{tabularx}
\caption{Supplied-group label matrices. Columns identify the second
member of each comparison; C/I/U mean correct/incorrect/unverifiable.}
\label{tab:human_group_confusion}
\end{table*}

Table~\ref{tab:human_score_intervals} compares automatic--H1 and independent
H1--H2 response-score agreement. In the primary automatic--H1 comparison,
both sides retain their recorded quantity rules: $N$ differs in 12
responses and count mode in 11 (15 responses in the union).
All 200 responses enter the comparison with their recorded scores. For the
two sampled model responses per task, automatic and human orderings agree on
56/60 tasks for AS and 57/61 for AP when both orderings are non-tied.
Including 28 shared AS ties and 27 shared AP ties gives 84/100 matching
relations for each metric. Each metric also has four reversed non-ties and
12 one-sided ties.

\begin{table*}[tbp]
\centering\suppTableSetup
\fontsize{9}{11}\selectfont
\setlength{\tabcolsep}{3pt}
\begin{tabularx}{\linewidth}{@{}L{0.10\linewidth}YYYYYY@{}}
\toprule
& \multicolumn{3}{c}{Automatic--H1} &
\multicolumn{3}{c}{H1--H2} \\
\cmidrule(lr){2-4}\cmidrule(lr){5-7}
Metric & AS & AP & Upper & AS & AP & Upper \\
\midrule
Pearson $r$ & \shortstack[l]{0.9054\\{\fontsize{8}{9}\selectfont[0.8496, 0.9519]}} & \shortstack[l]{0.9132\\{\fontsize{8}{9}\selectfont[0.8595, 0.9588]}} & \shortstack[l]{0.8798\\{\fontsize{8}{9}\selectfont[0.8107, 0.9372]}} & \shortstack[l]{0.8961\\{\fontsize{8}{9}\selectfont[0.7335, 0.9825]}} & \shortstack[l]{0.9022\\{\fontsize{8}{9}\selectfont[0.7468, 0.9834]}} & \shortstack[l]{0.9215\\{\fontsize{8}{9}\selectfont[0.8003, 0.9879]}} \\
Spearman $\rho$ & \shortstack[l]{0.9035\\{\fontsize{8}{9}\selectfont[0.8441, 0.9515]}} & \shortstack[l]{0.9103\\{\fontsize{8}{9}\selectfont[0.8514, 0.9583]}} & \shortstack[l]{0.8816\\{\fontsize{8}{9}\selectfont[0.8107, 0.9370]}} & \shortstack[l]{0.8614\\{\fontsize{8}{9}\selectfont[0.6248, 0.9837]}} & \shortstack[l]{0.8679\\{\fontsize{8}{9}\selectfont[0.6394, 0.9855]}} & \shortstack[l]{0.9003\\{\fontsize{8}{9}\selectfont[0.7314, 0.9876]}} \\
Kendall $\tau$ & \shortstack[l]{0.8609\\{\fontsize{8}{9}\selectfont[0.7931, 0.9204]}} & \shortstack[l]{0.8712\\{\fontsize{8}{9}\selectfont[0.8054, 0.9299]}} & \shortstack[l]{0.8326\\{\fontsize{8}{9}\selectfont[0.7557, 0.8992]}} & \shortstack[l]{0.8087\\{\fontsize{8}{9}\selectfont[0.5839, 0.9578]}} & \shortstack[l]{0.8136\\{\fontsize{8}{9}\selectfont[0.5899, 0.9623]}} & \shortstack[l]{0.8561\\{\fontsize{8}{9}\selectfont[0.6881, 0.9675]}} \\
MAE & \shortstack[l]{4.8792\\{\fontsize{8}{9}\selectfont[2.8749, 7.1625]}} & \shortstack[l]{4.4542\\{\fontsize{8}{9}\selectfont[2.4832, 6.6833]}} & \shortstack[l]{6.2625\\{\fontsize{8}{9}\selectfont[3.7458, 9.1167]}} & \shortstack[l]{5.0625\\{\fontsize{8}{9}\selectfont[1.5625, 9.3958]}} & \shortstack[l]{5.0625\\{\fontsize{8}{9}\selectfont[1.5625, 9.3958]}} & \shortstack[l]{4.5625\\{\fontsize{8}{9}\selectfont[1.3333, 8.7297]}} \\
Bias & \shortstack[l]{0.8375\\{\fontsize{8}{9}\selectfont[$-1.4584$, 3.2500]}} & \shortstack[l]{0.4625\\{\fontsize{8}{9}\selectfont[$-1.7584$, 2.7833]}} & \shortstack[l]{2.0542\\{\fontsize{8}{9}\selectfont[$-0.7502$, 5.1333]}} & \shortstack[l]{0.8958\\{\fontsize{8}{9}\selectfont[$-3.3333$, 5.7500]}} & \shortstack[l]{0.8958\\{\fontsize{8}{9}\selectfont[$-3.3333$, 5.7500]}} & \shortstack[l]{0.3958\\{\fontsize{8}{9}\selectfont[$-3.5417$, 5.0625]}} \\
\bottomrule
\end{tabularx}
\caption{Response-score agreement on 200 automatic--H1 responses and the
independent 40-response H1--H2 subset. Each cell gives the point estimate
above its 95\% confidence interval. AS is Answer Score, AP is Answer
Precision, and Upper is Upper Score. AS/Upper errors are points on a
0--100 scale; AP errors are percentage points and AP correlations use
0--1 values. Bias is automatic minus H1 in the left block and H1 minus H2
in the right block. H2 labels do not replace H1 in the primary comparison.}
\label{tab:human_score_intervals}
\label{tab:second_reviewer_scores}
\end{table*}

\subsection{Returned-Link Feature Correspondence}
Every retained link in the process sample is reviewed: 317 candidate
edges across 55 trajectories and 37 tasks. Support requires a verified,
locatable feature---such as an entity, URL, domain, model, provider, or
price---with evidence on both the round and target sides. For a split or
merged group, at least one explicitly mapped target must satisfy this rule.
Review confirms 311/317 links, giving
\textbf{98.11\% feature-correspondence precision} ($[96.15,99.40]$),
with six unsupported and no undetermined links.

Explicit group alignment yields 475 alignment records and 360 targets,
while the evaluation unit remains the 317 original candidate edges.
Three targets retain reviewed machine-only text rather than a matched human
group. The criterion applies to observable content features across process
verdicts. It measures the precision of returned locations for retrospective
inspection of answer-related content.

\subsection{Commitment-Based Process Audit}
The separate human process audit uses \textbf{error} for an explicit
erroneous interpretation or commitment. Opening a candidate, observing
third-party content, and collecting an incomplete set of fields are assessed
as exploratory activity unless an erroneous commitment is present.
Automatic flags instead assess candidate content against hard constraints.
Table~\ref{tab:human_process_confusion} reports these two annotation targets:
the commitment-based audit assigns zero error labels, while the automatic
candidate-content procedure flags 349 rounds.

\begin{table}[!htb]
\centering\suppTableSetup
\begin{tabularx}{\linewidth}{@{}Yrrrr@{}}
\toprule
Human $\backslash$ Automatic & Error & Non-error & Skipped & Total \\
\midrule
Error & 0 & 0 & 0 & 0 \\
Non-error & 329 & 589 & 353 & 1,271 \\
Skipped & 20 & 9 & 1,342 & 1,371 \\
\midrule
Total & 349 & 598 & 1,695 & 2,642 \\
\bottomrule
\end{tabularx}
\caption{Complete process cross-tabulation on 100 trajectories from
50 tasks. The two sets of labels concern different annotation targets.}
\label{tab:human_process_confusion}
\end{table}

On the independent process subset, H1/H2 agree on 494/499 rounds
(99.00\%, $[97.74,100.00]$; $\kappa=0.9799$, $[0.9525,1.0000]$).
The five disagreements are H1-\textbf{skipped}/H2-\textbf{non\_error}.
Both reviewers assign zero error labels; their agreement therefore describes
the distinction between non-error and skipped rounds. Binary-error $\kappa$
and conditional type agreement are undefined in this subset.

For the 317 candidate links, locked human process endpoints comprise
289 non-error and 28 skipped labels. A separate strict conjunction requires
the \emph{same target} to have a human error node, a locked incorrect answer
group, and supported feature correspondence. It yields 0/317 edges because
the reviewed process endpoints contain no human error label. The
311/317 feature-correspondence result and this conjunction retain their
distinct definitions and the same original edge denominator.


%% file: supplementary_sections/08b_resources.tex
\section{Computational Resources and API Budget}
\label{app:resource_costs}

Constructing and evaluating the corpus requires sustained model inference,
web interaction, and source verification across 1,600 retained runs.
The cumulative runtimes reported in the main paper are 1,037 hours for
trajectory collection and 318 hours for automatic evaluation, totaling
\textbf{1,355 hours}. The upper-scenario API budget is approximately
\textbf{USD 2,943}: USD 1,812.30 for collection and USD 1,130.69 for
evaluation. Runtime, interaction volume, and API workload together describe
the investment in collecting and verifying the retained corpus.

\subsection{Accounting Scope and Workload}
\label{app:resource_scope}

Runtime follows the main-paper totals and sums the time invested across
runs; parallel execution determines the elapsed calendar schedule.
API budgets use the archived price snapshot dated 23 October 2026.
Collection uses the upper full-history workload scenario, while evaluation
uses recorded token usage repriced under the paper's fixed evaluator roles.
All amounts are in USD and cover model API inference for the retained-run
pipeline. Web-search services, hosting, labor, and development experiments
are separate resource categories.

The collection records contain 74,015 tool-result events: 24,482 searches,
21,351 page openings, 15,302 reads, 12,193 within-page searches, and 687 link
extractions. These interactions expose 44.58 million read characters and
produce the 42,658 normalized rounds used in evaluation. Reconstructing the
model-request workload yields an estimated 75,622 calls. Long trajectories
repeatedly present accumulated context to the model, so input processing is
a substantial component of the collection budget.

Evaluation records contain 40,455 structured model operations: 30,335 in
Stage~1 and 10,120 in Stage~2. Recorded usage totals \textbf{422.56 million
tokens}, comprising 259.23 million input and 163.32 million output tokens.
The operation records also include 14,722 JSON-repair attempts, yielding at
least 55,177 successful provider responses. Their usage is already included
in the recorded token totals. Collection calls and evaluation operations
retain their respective counting units.

\subsection{Upper-Scenario Collection Budget}
\label{app:collection_budget}

The upper workload scenario reconstructs full accumulated input histories
and allocates 800 output tokens per model request. Saved histories use
\textbf{cl100k\_base} as a common token-count proxy, yielding 1,467,125,200
input tokens. For 72,913 tool-associated assistant-message slots,
the scenario allocates 530 visible tokens per
message; their repetition in subsequent requests contributes 2,482,483
history occurrences. Thus the aggregate input and output budgets are
\[
\begin{aligned}
T_{\mathrm{in}}^{\mathrm{col}}
 &= 1{,}467{,}125{,}200 + 530\times2{,}482{,}483\\
 &= 2{,}782{,}841{,}190,\\
T_{\mathrm{out}}^{\mathrm{col}}
 &= 75{,}622\times800 = 60{,}497{,}600.
\end{aligned}
\]
This is the upper full-history scenario in the resource audit. Each model's
input and output workloads are priced separately at its archived uncached
rates. The resulting collection budget is USD 1,812.30, of which input
processing contributes USD 1,624.44 (89.63\%) and output generation
USD 187.86. Each model contributes 100 retained trajectories. The largest
collection budgets are USD 334.61 for Gemini-3.1-Pro, USD 278.07 for
GPT-6.1-Sol, and USD 263.06 for GPT-5.4; GPT-6-Luna-Pro has the smallest
budget at USD 9.82. This variation reflects both accumulated-context workload
and the archived model-specific input/output rates.

\subsection{Evaluation Budget and Stage Allocation}
\label{app:evaluation_budget}

For role $r$, let $T_r^{\mathrm{in}}$ and $T_r^{\mathrm{out}}$ be recorded
token counts, and $p_r^{\mathrm{in}}$ and $p_r^{\mathrm{out}}$ the reference
prices in USD per million tokens. The budget is
\[
B_{\mathrm{API}}=
\sum_r\frac{T_r^{\mathrm{in}}p_r^{\mathrm{in}}
                 +T_r^{\mathrm{out}}p_r^{\mathrm{out}}}{10^6}.
\]
The fixed roles use DeepSeek-V4-Pro for Stage~1 and Judge~A, with input/output
rates of USD 1.32/3.96 per million tokens. Judge~B and the arbiter use
Qwen3.7-Max at USD 2.50/7.50. These are the audit's peak/international
reference rates with uncached input.

Stage~1 comprises rubric construction, answer grouping, source verification,
and recorded evidence-repair operations. Its budget is USD 676.66 (59.84\%
of evaluation). Stage~2's two judges and arbitration account for USD 454.03
(40.16\%). Table~\ref{tab:resource_budget} summarizes the token workloads
and their corresponding upper-scenario budgets.

\begin{table}[!htb]
\centering
\suppTableSetup
\begin{tabularx}{\linewidth}{@{}Yrrr@{}}
\toprule
\textbf{Phase} & \textbf{Input (M)} & \textbf{Output (M)} &
\textbf{USD} \\
\midrule
Collection & 2,782.84 & 60.50 & 1,812.30 \\
Stage~1 & 157.17 & 118.48 & 676.66 \\
Stage~2 & 102.07 & 44.84 & 454.03 \\
\midrule
\textbf{Budgeted total} & \textbf{3,042.07} & \textbf{223.82} &
\textbf{2,942.98} \\
\bottomrule
\end{tabularx}
\caption{Token workload and upper-scenario API budget for the retained-run
pipeline. M denotes one million tokens. Collection uses estimated workload;
evaluation uses recorded usage. All API amounts are reference-price budgets.
Totals are computed before rounding.}
\label{tab:resource_budget}
\end{table}

Together, the runtime and workload breakdowns quantify the investment in
collecting real interactions and verifying their outcomes. The budget includes
repeated context processing, source-constrained verification, independent
round judgments, and arbitration as distinct parts of the evaluation pipeline.
The resource audit records the selected workload scenario,
price snapshot, and stage allocation for offline re-aggregation.

%% file: supplementary_sections/11_reproduction.tex
\section{Prompts, Schemas, Algorithms, and Reproduction}
\label{app:reproduction}

This section specifies the retained prompt and schema contracts, implementation
procedure, and three reproduction modes.

\subsection{Prompt Roles and Shared Information Boundaries}
\label{app:prompt_inventory}

The seven prompt roles are the collection agent, three Stage~1 roles
(rubric construction, extraction, verification), two Stage~2 judges, and the
arbiter. Evaluator inputs exclude evaluated-model identity. Stage~1 accesses
agent-cited URLs; Stage~2 uses the recorded audit tuple and arbitration adds
both judge outputs.

\subsection{Agent System Prompt and Tool Descriptions}
\label{app:agent_prompt}

The shared instruction defines the ReAct syntax, evidence-before-answer
rule, and terminal format. The archived instruction is reproduced below with its wording unchanged:

\begin{promptbox}{Agent instruction (archived source snapshot)}
\normalsize\setlength{\parindent}{0pt}\setlength{\parskip}{4pt}
You are a web research ReAct agent in a weak-tool evaluation setting.

Every task requires web search. Use web\_search to discover candidate URLs, but web\_search returns only titles and URLs.

You must open and inspect pages using open\_url, read\_page, find\_on\_page, and extract\_links.

Do not answer from search result titles alone. Do not assume a page supports a claim unless you have read relevant page text.

Use multiple pages when the task requires comparison or verification.

When calling a tool, use exactly one tool call and no extra text after the Action Input. Use this format:\\
Thought: ...\\
Action: tool\_name\\
Action Input: \{"arg": "value"\}

Never output only a JSON object or a fenced JSON block. The tool name must always appear on an Action line before Action Input.

Never output provider-style tool text such as [Tool call], function\_call, function\_calls, or standalone tool names followed by JSON. Those are not valid final answers and not valid ReAct steps in this experiment.

Do not output chat role labels such as User:, Assistant:, or System:.

Do not use Tool or Tool Input blocks. Do not use XML-style, angle-bracket, or provider-specific function-call markup. Do not place a new Thought or Action on the same line as Action Input.

When you have enough evidence, use the official ReAct final format:\\
Thought: I can answer without using any more tools.\\
Answer: [final answer with source URLs]

After Answer, stop completely. Never request read\_page, web\_search, open\_url, find\_on\_page, or extract\_links after Answer.

Do not use Answer to explain that tools failed or that evidence is missing. A final Answer must directly satisfy the user's task using verified source URLs. If evidence is insufficient, continue using the available tools instead of ending with a refusal or noncompletion note.

Do not use Action after Answer.
\end{promptbox}

The controller appends the five tool descriptions and arguments in
Appendix~\ref{app:weak_tool_api}.

\subsection{Stage~1 Prompt Contracts}
\label{app:stage1_prompts}

\paragraph{Rubric construction.}
The task yields $N$, \textbf{count\_mode}, hard constraints, and a grouping
rule, preserving strict inequalities, dates, scope, source requirements, and
conditional applicability. Count requirements govern task-level scoring;
other constraints govern eligible groups.

\paragraph{Answer-group extraction.}
The final response yields ordered task-defined units with stable IDs and
associated URLs. Extraction resolves formatting while preserving answer
content and cited sources.

\paragraph{Group verification.}
The verifier uses each group, its usable cited texts, and answer-visible
requirements to return a verdict, failed criterion IDs, evidence references, inaccessible URLs where
applicable, and an evidence-grounded reason. The access-only rule follows
Appendix~\ref{app:stage1_group_verdicts}.

\subsection{Stage~2 Judge and Arbiter Prompt Contracts}
\label{app:stage2_prompts}

Each judge independently maps the tuple in
Appendix~\ref{app:stage2_audit_inputs} to a verdict, failed criteria,
a diagnostic type, proposed group IDs, and a reason using the candidate-content
checks. Links use observable entity, URL, date, price, or other task-field
correspondences. The arbiter receives the tuple plus both judge records and
returns the same round-evaluation schema for verdict disagreements.

\paragraph{Source bindings.}
The archived implementation snapshot records prompt builders and schemas under
\path{method_artifacts/prompts_and_schemas/}:
\path{agent_system_prompt_current.txt};
\path{prompt_and_tool_builder_inventory.json} (source symbols and verbatim
builder bodies); and \path{schemas_current.py} (typed schema). These specify
snapshot wording and serialization. Retained manifests govern historical
execution settings; rendered provider requests and per-run environment
overrides have partial archival coverage.

\subsection{Structured Output Schemas}
\label{app:json_schemas}

Machine-readable schemas specify exact types, optionality, enumerations, and
validation rules for four object groups:

\suppEntry{Collection objects}{
Run/task IDs, question, final response, ordered rounds, and model metadata;
rounds contain IDs, model text, tool sequences, observations, URLs, and
page-text context.
}
\suppEntry{Stage~1 objects}{
Rubric fields above plus criterion IDs; group content, citations, and access
states; evaluation verdicts, criterion results, evidence references, reasons,
counts, and scores.
}
\suppEntry{Stage~2 objects}{
Round ID, verdict, skip flag, failed criteria, type, links, and reason;
consensus adds judge verdicts, arbiter-use status, and merged or
arbiter-returned links before target-status filtering.
}
\suppEntry{Final report}{
Stage outputs joined to linkage summaries, behavioral fields, and execution
metadata by retained run IDs.
}

\subsection{Implementation-Aligned Procedures}
\label{app:reproduction_algorithms}

The implementation applies the Stage~2 rules in
Appendix~\ref{app:stage2_localization} in the
following order, preserving proposed links before target-status filtering.

\begingroup
\setlist[enumerate]{leftmargin=*,itemsep=0pt,parsep=0pt,topsep=2pt}
\begin{suppalgorithm}{Dual-judge consensus and downstream linkage summaries}
\label{alg:stage2_consensus}
\begin{enumerate}
    \item Pre-skip structurally empty rounds and configured background or
    navigation-pattern matches.
    \item Send remaining rounds independently to both judges with identical
    audit tuples and output contracts.
    \item On error agreement, retain A's base record, union failed criteria,
    deduplicate the linked-ID union while preserving first occurrences, and use A's normalized
    type or, if unavailable, B's.
    \item When both judges agree on nonerror or skipped, retain A's record; on verdict disagreement,
    retain the arbiter record. Type disagreement alone does not invoke arbitration.
    \item Persist evaluable records and compact skipped-round aggregates.
    \item Normalize only stored ID shape and retain
    \(\mathcal{L}^{\mathrm{sum}}_t=\bar{\mathcal{L}}_t\cap\mathcal{B}\), where
    \(t\) is a consensus error node and \(\mathcal{B}\) contains Stage~1-incorrect
    group IDs.
    \item Aggregate total, evaluable, error, typed, linked, and first-linked
    diagnostics; links record diagnostic content associations.
\end{enumerate}
\end{suppalgorithm}
\endgroup

\Needspace{4\baselineskip}
\subsection{Artifact Scope and Reproduction}
\label{app:release_inventory}\label{app:reproduction_pipeline}
\label{app:reaggregation_boundary}
The anonymized repository provides core implementation, representative samples,
and instructions identifying released materials and executable entry points.
Credentials and private provider settings remain outside the public artifacts.

\suppEntry{Offline re-aggregation}{
Join the 1,600 retained manifests and trajectories to tasks, Stage~1 rubrics,
groups and verdicts, Stage~2 judge and consensus records, and behavioral
and aggregation tables. Use the frozen run-selection manifest and task/run
IDs, preserving one row per model--task pair, its scoring mode, and
target-status filtering. All records come from the same frozen version.
}
\suppEntry{Evaluator re-execution}{
Re-running either stage calls external providers under selected configurations.
Provider revisions, prompt versions, serialization, and serving nondeterminism
may alter outputs.
}
\suppEntry{Live-web re-verification}{
Stage~1 revisits agent-cited URLs under current access conditions: pages may
change, redirect, disappear, or become blocked. Released page excerpts and
report references preserve selected observations; live retrieval produces new ones.
}

%% file: supplementary_sections/12_limitations.tex
\section{Evaluation Scope and Broader Impacts}
\label{app:limitations}

\subsection{Task and Execution Scope}
\label{app:limitation_task_scope}
\label{app:limitation_weak_tools}
The 100 public-web tasks evaluate multi-source search and constraint
satisfaction through source-supported answer groups under a shared weak-tool
interface.
One retained run per model--task cell supports paired task comparisons;
reported uncertainty concerns task variation. Visual interaction,
authenticated workflows, private context, transactions, and richer browser
operations define complementary settings with different access and execution
policies.

\subsection{Evaluator Configuration and Rubric Variation}
\label{app:limitation_evaluators}
\label{app:limitation_rubrics}
Fixed evaluator roles standardize comparisons. Human-agreement statistics use separate primary-sample and
independent-subset denominators
(Appendix~\ref{app:evaluator_reliability}). Required counts are more consistent
across run-specific rubrics than hard-constraint counts; field-level modal
coverage and normalized text hashes characterize complementary aspects
(Appendix~\ref{app:rubric_stability}). Other judges, prompts, or provider
versions provide configurations for sensitivity analysis.

\subsection{Process Interpretation}
\label{app:limitation_linking_taxonomy}
\label{app:limitation_observable_process}
Link review measures returned feature correspondence; commitment review
assesses explicit erroneous commitments (Appendix~\ref{app:evaluator_reliability}).
Stage~2 conditions on final responses and Stage~1 verdicts, so flag--score
associations are outcome-conditioned. The earliest linked-round index locates the first
returned association (Appendix~\ref{app:stage2_diagnostics}); the taxonomy
assigns one dominant category from ten labels and accounts separately for
unassigned types.

\subsection{Replication Conditions}
\label{app:limitation_web_drift}
Appendix~\ref{app:reaggregation_boundary} distinguishes frozen-record
aggregation from evaluator re-execution and live-web verification under new
provider and source conditions. The budget snapshot fixes reference prices
and upper workload assumptions (Appendix~\ref{app:resource_costs}).

\subsection{Broader Impacts}
\label{app:broader_impacts}
Source-supported scores and inspectable trajectory associations support
evaluation research and targeted agent improvement within this setting.
Public-web excerpts require attention to licenses, privacy, and redistribution;
improved browsing also enables large-scale scraping. Security, prompt-injection
resilience, and deployment safety require dedicated evaluations. Offline
aggregation reuses collected model and network outputs; energy use and
emissions lie outside the reported resource accounting.

%% file: references.bib
@inproceedings{deng2023mind2web,
  author    = {Xiang Deng and Yu Gu and Boyuan Zheng and Shijie Chen and Samuel Stevens and Boshi Wang and Huan Sun and Yu Su},
  title     = {{Mind2Web}: Towards a Generalist Agent for the Web},
  booktitle = {Advances in Neural Information Processing Systems},
  year      = {2023},
  volume    = {36}
}

@inproceedings{liu2024agentbench,
  author    = {Xiao Liu and Hao Yu and Hanchen Zhang and Yifan Xu and Xuanyu Lei and Hanyu Lai and Yu Gu and Hangliang Ding and Kaiwen Men and Kejuan Yang and Shudan Zhang and Xiang Deng and Aohan Zeng and Zhengxiao Du and Chenhui Zhang and Sheng Shen and Tianjun Zhang and Yu Su and Huan Sun and Minlie Huang and Yuxiao Dong and Jie Tang},
  title     = {{AgentBench}: Evaluating {LLM}s as Agents},
  booktitle = {The Twelfth International Conference on Learning Representations},
  year      = {2024}
}

@inproceedings{zhou2024webarena,
  author    = {Shuyan Zhou and Frank F. Xu and Hao Zhu and Xuhui Zhou and Robert Lo and Abishek Sridhar and Xianyi Cheng and Tianyue Ou and Yonatan Bisk and Daniel Fried and Uri Alon and Graham Neubig},
  title     = {{WebArena}: A Realistic Web Environment for Building Autonomous Agents},
  booktitle = {The Twelfth International Conference on Learning Representations},
  year      = {2024}
}

@inproceedings{he2024webvoyager,
  author    = {Hongliang He and Wenlin Yao and Kaixin Ma and Wenhao Yu and Yong Dai and Hongming Zhang and Zhenzhong Lan and Dong Yu},
  title     = {{WebVoyager}: Building an End-to-End Web Agent with Large Multimodal Models},
  booktitle = {Proceedings of the 62nd Annual Meeting of the Association for Computational Linguistics (Volume 1: Long Papers)},
  pages     = {6864--6890},
  year      = {2024}
}

@inproceedings{mialon2024gaia,
  author    = {Gr{\'e}goire Mialon and Cl{\'e}mentine Fourrier and Craig Swift and Thomas Wolf and Yann LeCun and Thomas Scialom},
  title     = {{GAIA}: A Benchmark for General {AI} Assistants},
  booktitle = {The Twelfth International Conference on Learning Representations},
  year      = {2024}
}

@inproceedings{zheng2023judging,
  author    = {Lianmin Zheng and Wei-Lin Chiang and Ying Sheng and Siyuan Zhuang and Zhanghao Wu and Yonghao Zhuang and Zi Lin and Zhuohan Li and Dacheng Li and Eric P. Xing and Hao Zhang and Joseph E. Gonzalez and Ion Stoica},
  title     = {Judging {LLM}-as-a-Judge with {MT-Bench} and Chatbot Arena},
  booktitle = {Advances in Neural Information Processing Systems},
  year      = {2023},
  volume    = {36}
}

@inproceedings{zeng2024llmbar,
  author    = {Zhiyuan Zeng and Jiatong Yu and Tianyu Gao and Yu Meng and Tanya Goyal and Danqi Chen},
  title     = {Evaluating Large Language Models at Evaluating Instruction Following},
  booktitle = {The Twelfth International Conference on Learning Representations},
  year      = {2024}
}

@inproceedings{ke2024critiquellm,
  author    = {Pei Ke and Bosi Wen and Andrew Feng and Xiao Liu and Xuanyu Lei and Jiale Cheng and Shengyuan Wang and Aohan Zeng and Yuxiao Dong and Hongning Wang and Jie Tang and Minlie Huang},
  title     = {{CritiqueLLM}: Towards an Informative Critique Generation Model for Evaluation of Large Language Model Generation},
  booktitle = {Proceedings of the 62nd Annual Meeting of the Association for Computational Linguistics (Volume 1: Long Papers)},
  pages     = {13034--13054},
  year      = {2024}
}

@inproceedings{niu2024ragtruth,
  author    = {Cheng Niu and Yuanhao Wu and Juno Zhu and Siliang Xu and KaShun Shum and Randy Zhong and Juntong Song and Tong Zhang},
  title     = {{RAGTruth}: A Hallucination Corpus for Developing Trustworthy Retrieval-Augmented Language Models},
  booktitle = {Proceedings of the 62nd Annual Meeting of the Association for Computational Linguistics (Volume 1: Long Papers)},
  pages     = {10862--10878},
  year      = {2024}
}

@inproceedings{lightman2024verify,
  author    = {Hunter Lightman and Vineet Kosaraju and Yuri Burda and Harrison Edwards and Bowen Baker and Teddy Lee and Jan Leike and John Schulman and Ilya Sutskever and Karl Cobbe},
  title     = {Let's Verify Step by Step},
  booktitle = {The Twelfth International Conference on Learning Representations},
  year      = {2024}
}

@inproceedings{zheng2025processbench,
  author    = {Chujie Zheng and Zhenru Zhang and Beichen Zhang and Runji Lin and Keming Lu and Bowen Yu and Dayiheng Liu and Jingren Zhou and Junyang Lin},
  title     = {{ProcessBench}: Identifying Process Errors in Mathematical Reasoning},
  booktitle = {Proceedings of the 63rd Annual Meeting of the Association for Computational Linguistics (Volume 1: Long Papers)},
  pages     = {1009--1024},
  doi       = {10.18653/v1/2025.acl-long.50},
  year      = {2025}
}

@inproceedings{yoran2024assistantbench,
  author    = {Ori Yoran and Samuel Joseph Amouyal and Chaitanya Malaviya and Ben Bogin and Ofir Press and Jonathan Berant},
  title     = {{AssistantBench}: Can Web Agents Solve Realistic and Time-Consuming Tasks?},
  booktitle = {Proceedings of the 2024 Conference on Empirical Methods in Natural Language Processing},
  pages     = {8938--8968},
  year      = {2024}
}

@misc{wang2025liveresearchbench,
  author        = {Jiayu Wang and Yifei Ming and Riya Dulepet and Qinglin Chen and Austin Xu and Zixuan Ke and Frederic Sala and Aws Albarghouthi and Caiming Xiong and Shafiq Joty},
  title         = {{LiveResearchBench}: A Live Benchmark for User-Centric Deep Research in the Wild},
  year          = {2025},
  eprint        = {2510.14240},
  archivePrefix = {arXiv}
}

@inproceedings{gou2025mind2web2,
  author    = {Boyu Gou and Zanming Huang and Yuting Ning and Yu Gu and Michael Lin and Weijian Qi and Andrei Kopanev and Botao Yu and Bernal Jim{\'e}nez Guti{\'e}rrez and Yiheng Shu and Chan Hee Song and Jiaman Wu and Shijie Chen and Hanane Moussa and Tianshu Zhang and Jian Xie and Yifei Li and Tianci Xue and Zeyi Liao and Kai Zhang and Boyuan Zheng and Zhaowei Cai and Viktor Rozgic and Morteza Ziyadi and Huan Sun and Yu Su},
  title     = {{Mind2Web 2}: Evaluating Agentic Search with Agent-as-a-Judge},
  booktitle = {Advances in Neural Information Processing Systems},
  volume    = {38},
  year      = {2025}
}

@misc{wang2026telbench,
  author        = {Jiaming Wang and Ziteng Feng and Jiangtao Wu and Ruihao Li and Qianqian Xie and Yuxiang Ren and He Zhu and Xueming Han and Fanyu Meng and Junlan Feng and Jiaheng Liu},
  title         = {Where Do Deep-Research Agents Go Wrong? Span-Level Error Localization in Agent Trajectories},
  year          = {2026},
  eprint        = {2606.02060},
  archivePrefix = {arXiv}
}

@inproceedings{zhang2024snowballing,
  author    = {Muru Zhang and Ofir Press and William Merrill and Alisa Liu and Noah A. Smith},
  title     = {How Language Model Hallucinations Can Snowball},
  booktitle = {Proceedings of the 41st International Conference on Machine Learning},
  pages     = {59670--59684},
  year      = {2024},
  volume    = {235},
  series    = {Proceedings of Machine Learning Research}
}

@inproceedings{lu2024weblinx,
  author    = {Xing Han Lu and Zden{\v{e}}k Kasner and Siva Reddy},
  title     = {{WebLINX}: Real-World Website Navigation with Multi-Turn Dialogue},
  booktitle = {Proceedings of the 41st International Conference on Machine Learning},
  pages     = {33007--33056},
  year      = {2024},
  volume    = {235},
  series    = {Proceedings of Machine Learning Research}
}

@inproceedings{koh2024visualwebarena,
  author    = {Jing Yu Koh and Robert Lo and Lawrence Jang and Vikram Duvvur and Ming Lim and Po-Yu Huang and Graham Neubig and Shuyan Zhou and Russ Salakhutdinov and Daniel Fried},
  title     = {{VisualWebArena}: Evaluating Multimodal Agents on Realistic Visual Web Tasks},
  booktitle = {Proceedings of the 62nd Annual Meeting of the Association for Computational Linguistics (Volume 1: Long Papers)},
  pages     = {881--905},
  year      = {2024}
}

@inproceedings{boisvert2024workarenaplus,
  author    = {L{\'e}o Boisvert and Megh Thakkar and Maxime Gasse and Massimo Caccia and Thibault Le Sellier De Chezelles and Quentin Cappart and Nicolas Chapados and Alexandre Lacoste and Alexandre Drouin},
  title     = {{WorkArena++}: Towards Compositional Planning and Reasoning-Based Common Knowledge Work Tasks},
  booktitle = {Advances in Neural Information Processing Systems},
  year      = {2024},
  volume    = {37}
}

@inproceedings{xie2024travelplanner,
  author    = {Jian Xie and Kai Zhang and Jiangjie Chen and Tinghui Zhu and Renze Lou and Yuandong Tian and Yanghua Xiao and Yu Su},
  title     = {{TravelPlanner}: A Benchmark for Real-World Planning with Language Agents},
  booktitle = {Proceedings of the 41st International Conference on Machine Learning},
  pages     = {54590--54613},
  year      = {2024},
  volume    = {235},
  series    = {Proceedings of Machine Learning Research}
}

@inproceedings{ma2024agentboard,
  author    = {Chang Ma and Junlei Zhang and Zhihao Zhu and Cheng Yang and Yujiu Yang and Yaohui Jin and Zhenzhong Lan and Lingpeng Kong and Junxian He},
  title     = {{AgentBoard}: An Analytical Evaluation Board of Multi-Turn {LLM} Agents},
  booktitle = {Advances in Neural Information Processing Systems},
  year      = {2024},
  volume    = {37}
}

@inproceedings{ru2024ragchecker,
  author    = {Dongyu Ru and Lin Qiu and Xiangkun Hu and Tianhang Zhang and Peng Shi and Shuaichen Chang and Cheng Jiayang and Cunxiang Wang and Shichao Sun and Huanyu Li and Zizhao Zhang and Binjie Wang and Jiarong Jiang and Tong He and Zhiguo Wang and Pengfei Liu and Yue Zhang and Zheng Zhang},
  title     = {{RAGChecker}: A Fine-Grained Framework for Diagnosing Retrieval-Augmented Generation},
  booktitle = {Advances in Neural Information Processing Systems},
  year      = {2024},
  volume    = {37}
}

@inproceedings{wang2024mint,
  author    = {Xingyao Wang and Zihan Wang and Jiateng Liu and Yangyi Chen and Lifan Yuan and Hao Peng and Heng Ji},
  title     = {{MINT}: Evaluating {LLM}s in Multi-turn Interaction with Tools and Language Feedback},
  booktitle = {The Twelfth International Conference on Learning Representations},
  year      = {2024}
}

@inproceedings{kim2024prometheus,
  author    = {Seungone Kim and Jamin Shin and Yejin Cho and Joel Jang and Shayne Longpre and Hwaran Lee and Sangdoo Yun and Seongjin Shin and Sungdong Kim and James Thorne and Minjoon Seo},
  title     = {{Prometheus}: Inducing Fine-Grained Evaluation Capability in Language Models},
  booktitle = {The Twelfth International Conference on Learning Representations},
  year      = {2024}
}

@inproceedings{trivedi2024appworld,
  author    = {Harsh Trivedi and Tushar Khot and Mareike Hartmann and Ruskin Manku and Vinty Dong and Edward Li and Shashank Gupta and Ashish Sabharwal and Niranjan Balasubramanian},
  title     = {{AppWorld}: A Controllable World of Apps and People for Benchmarking Interactive Coding Agents},
  booktitle = {Proceedings of the 62nd Annual Meeting of the Association for Computational Linguistics (Volume 1: Long Papers)},
  pages     = {16022--16076},
  year      = {2024}
}

@inproceedings{xie2024osworld,
  author    = {Tianbao Xie and Danyang Zhang and Jixuan Chen and Xiaochuan Li and Siheng Zhao and Ruisheng Cao and Toh Jing Hua and Zhoujun Cheng and Dongchan Shin and Fangyu Lei and Yitao Liu and Yiheng Xu and Shuyan Zhou and Silvio Savarese and Caiming Xiong and Victor Zhong and Tao Yu},
  title     = {{OSWorld}: Benchmarking Multimodal Agents for Open-Ended Tasks in Real Computer Environments},
  booktitle = {Advances in Neural Information Processing Systems},
  volume    = {37},
  year      = {2024}
}

@inproceedings{rawles2025androidworld,
  author    = {Chris Rawles and Sarah Clinckemaillie and Yifan Chang and Jonathan Waltz and Gabrielle Lau and Marybeth Fair and Alice Li and William Bishop and Wei Li and Folawiyo Campbell-Ajala and Daniel Toyama and Robert Berry and Divya Tyamagundlu and Timothy Lillicrap and Oriana Riva},
  title     = {{AndroidWorld}: A Dynamic Benchmarking Environment for Autonomous Agents},
  booktitle = {The Thirteenth International Conference on Learning Representations},
  year      = {2025}
}

@inproceedings{tan2025judgebench,
  author    = {Sijun Tan and Siyuan Zhuang and Kyle Montgomery and William Tang and Alejandro Cuadron and Chenguang Wang and Raluca Popa and Ion Stoica},
  title     = {{JudgeBench}: A Benchmark for Evaluating {LLM}-Based Judges},
  booktitle = {The Thirteenth International Conference on Learning Representations},
  year      = {2025}
}

@inproceedings{wang2024unfairevaluators,
  author    = {Peiyi Wang and Lei Li and Liang Chen and Zefan Cai and Dawei Zhu and Binghuai Lin and Yunbo Cao and Lingpeng Kong and Qi Liu and Tianyu Liu and Zhifang Sui},
  title     = {Large Language Models Are Not Fair Evaluators},
  booktitle = {Proceedings of the 62nd Annual Meeting of the Association for Computational Linguistics (Volume 1: Long Papers)},
  pages     = {9440--9450},
  year      = {2024}
}

@inproceedings{panickssery2024selfpreference,
  author    = {Arjun Panickssery and Samuel R. Bowman and Shi Feng},
  title     = {{LLM} Evaluators Recognize and Favor Their Own Generations},
  booktitle = {Advances in Neural Information Processing Systems},
  volume    = {37},
  year      = {2024}
}

@inproceedings{wang2024mathshepherd,
  author    = {Peiyi Wang and Lei Li and Zhihong Shao and Runxin Xu and Damai Dai and Yifei Li and Deli Chen and Yu Wu and Zhifang Sui},
  title     = {Math-Shepherd: Verify and Reinforce {LLM}s Step-by-Step without Human Annotations},
  booktitle = {Proceedings of the 62nd Annual Meeting of the Association for Computational Linguistics (Volume 1: Long Papers)},
  pages     = {9426--9439},
  year      = {2024}
}

@inproceedings{liu2024hdeval,
  author    = {Yuxuan Liu and Tianchi Yang and Shaohan Huang and Zihan Zhang and Haizhen Huang and Furu Wei and Weiwei Deng and Feng Sun and Qi Zhang},
  title     = {{HD-Eval}: Aligning Large Language Model Evaluators through Hierarchical Criteria Decomposition},
  booktitle = {Proceedings of the 62nd Annual Meeting of the Association for Computational Linguistics (Volume 1: Long Papers)},
  pages     = {7641--7660},
  year      = {2024}
}

@inproceedings{yao2023react,
  author    = {Shunyu Yao and Jeffrey Zhao and Dian Yu and Nan Du and Izhak Shafran and Karthik Narasimhan and Yuan Cao},
  title     = {{ReAct}: Synergizing Reasoning and Acting in Language Models},
  booktitle = {The Eleventh International Conference on Learning Representations},
  year      = {2023}
}

@article{nakano2021webgpt,
  author    = {Reiichiro Nakano and Jacob Hilton and Suchir Balaji and Jeff Wu and Long Ouyang and Christina Kim and Christopher Hesse and Shantanu Jain and Vineet Kosaraju and William Saunders and Xu Jiang and Karl Cobbe and Tyna Eloundou and Gretchen Krueger and Kevin Button and Matthew Knight and Benjamin Chess and John Schulman},
  title     = {{WebGPT}: Browser-Assisted Question-Answering with Human Feedback},
  journal   = {arXiv preprint arXiv:2112.09332},
  year      = {2021}
}

@inproceedings{drouin2024workarena,
  author    = {Alexandre Drouin and Maxime Gasse and Massimo Caccia and Issam H. Laradji and Del Verme, Manuel and Tom Marty and David Vazquez and Nicolas Chapados and Alexandre Lacoste},
  title     = {{WorkArena}: How Capable are Web Agents at Solving Common Knowledge Work Tasks?},
  booktitle = {Proceedings of the 41st International Conference on Machine Learning},
  series    = {Proceedings of Machine Learning Research},
  volume    = {235},
  pages     = {11642--11662},
  year      = {2024}
}

@inproceedings{wei2024longformfactuality,
  author    = {Jerry Wei and Chengrun Yang and Xinying Song and Yifeng Lu and Nathan Hu and Jie Huang and Dustin Tran and Daiyi Peng and Ruibo Liu and Da Huang and Cosmo Du and Quoc V. Le},
  title     = {Long-form Factuality in Large Language Models},
  booktitle = {Advances in Neural Information Processing Systems},
  volume    = {37},
  year      = {2024}
}

@article{wei2025browsecomp,
  author    = {Jason Wei and Zhiqing Sun and Spencer Papay and Scott McKinney and Jeffrey Han and Isa Fulford and Hyung Won Chung and Alex Tachard Passos and William Fedus and Amelia Glaese},
  title     = {{BrowseComp}: A Simple Yet Challenging Benchmark for Browsing Agents},
  journal   = {arXiv preprint arXiv:2504.12516},
  year      = {2025}
}

@inproceedings{yao2025taubench,
  author    = {Shunyu Yao and Noah Shinn and Pedram Razavi and Karthik Narasimhan},
  title     = {$\tau$-bench: A Benchmark for Tool-Agent-User Interaction in Real-World Domains},
  booktitle = {The Thirteenth International Conference on Learning Representations},
  year      = {2025}
}

@inproceedings{brown2020language,
  author    = {Tom B. Brown and Benjamin Mann and Nick Ryder and Melanie Subbiah and Jared D. Kaplan and Prafulla Dhariwal and Arvind Neelakantan and Pranav Shyam and Girish Sastry and Amanda Askell and Sandhini Agarwal and Ariel Herbert-Voss and Gretchen Krueger and Tom Henighan and Rewon Child and Aditya Ramesh and Daniel Ziegler and Jeffrey Wu and Clemens Winter and Chris Hesse and Mark Chen and Eric Sigler and Mateusz Litwin and Scott Gray and Benjamin Chess and Jack Clark and Christopher Berner and Sam McCandlish and Alec Radford and Ilya Sutskever and Dario Amodei},
  title     = {Language Models are Few-Shot Learners},
  booktitle = {Advances in Neural Information Processing Systems},
  volume    = {33},
  year      = {2020}
}
